\documentclass[twoside]{article}

\usepackage[accepted]{aistats2019}

\usepackage{time}
\RequirePackage{etex} 

\usepackage[round]{natbib}

\usepackage[utf8]{inputenc} %
\usepackage[T1]{fontenc}    %
\usepackage{url}            %
\usepackage{booktabs}       %
\usepackage{amsfonts,amsthm}       %
\usepackage{color, colortbl}
\usepackage[dvipsnames]{xcolor}

\usepackage{amsmath,amsfonts,bm}
\usepackage{algorithm,algorithmic}

\def\1{\bm{1}}

\def\vtheta{{\bm{\theta}}}
\def\vphi{{\bm{\varphi}}}

\def\vomega{{\bm{\omega}}}

\def\vb{{\bm{b}}}
\def\vc{{\bm{c}}}

\def\vu{{\bm{u}}}
\def\vv{{\bm{v}}}

\def\mA{{\bm{A}}}

\def\mD{{\bm{D}}}

\def\mI{{\bm{I}}}

\def\mM{{\bm{M}}}

\def\mP{{\bm{P}}}

\def\mU{{\bm{U}}}
\def\mV{{\bm{V}}}

\DeclareMathAlphabet{\mathsfit}{\encodingdefault}{\sfdefault}{m}{sl}
\SetMathAlphabet{\mathsfit}{bold}{\encodingdefault}{\sfdefault}{bx}{n}

\newcommand{\R}{\mathbb{R}}

\DeclareMathOperator*{\argmin}{arg\,min}

\DeclareMathOperator{\diag}{diag}

\newlength\myindent
\definecolor{beaublue}{rgb}{0.74, 0.83, 0.9}
\definecolor{babyblueeyes}{rgb}{0.63, 0.79, 0.95}
\definecolor{mydarkblue}{rgb}{0,0.08,0.45}
\usepackage[colorlinks=true,
    linkcolor=mydarkblue,
    citecolor=mydarkblue,
    filecolor=mydarkblue,
    urlcolor=mydarkblue]{hyperref}  
\definecolor{bettergreen}{HTML}{266A2E}
\definecolor{myblue}{HTML}{D2E4FC}

\newcommand{\green}{\color{bettergreen}}
\newcommand{\red}{\color{red}}

\DeclareMathOperator*{\Sp}{Sp} 

\newcommand{\bigzero}{\mbox{\normalfont\Large\bfseries 0}}

\usepackage{pifont}
\newcommand{\cmark}{\ding{51}}%
\newcommand{\xmark}{\ding{55}}%

\usepackage{todonotes}

\def\vv{{\boldsymbol v}}

\def\LL{{\mathcal L}}

\def\defas{:=}

\newtheorem{theorem}{Theorem}
\newtheorem{lemma}{Lemma}
\newtheorem{proposition}{Proposition}

\newtheorem*{rep@theorem}{\rep@title}
\newcommand{\newreptheorem}[2]{%
\newenvironment{rep#1}[1]{%
 \def\rep@title{#2 \ref{##1}}%
 \begin{rep@theorem}}%
 {\end{rep@theorem}}}
\newreptheorem{theorem}{Theorem'}
\newtheorem*{rep@proposition}{\rep@title}
\newcommand{\newrepproposition}[2]{%
\newenvironment{rep#1}[1]{%
 \def\rep@title{#2 \ref{##1}}%
 \begin{rep@proposition}}%
 {\end{rep@proposition}}}
\newreptheorem{proposition}{Proposition'}
\usepackage{nicefrac}       %
\usepackage{microtype}      %
\usepackage{graphicx}
\usepackage{subfigure}
\usepackage{tikz}
\usepackage{mathdots}
\usepackage{yhmath}
\usepackage{cancel}
\usepackage{blindtext}
\usepackage{siunitx}
\usepackage{array}
\usepackage{commath}
\usepackage{mdframed}
\usepackage{enumerate}
\usepackage{bold-extra}
\usepackage{nccmath}
\usepackage{caption}
\usepackage{dsfont}
\usepackage{gensymb}
\usepackage{tabularx}
\usepackage{amsmath, amsthm, amssymb}
\usepackage{enumitem}
\usepackage{booktabs}
\usepackage{multirow}
\usepackage{rotating}
\usepackage{mathtools}
\usepackage{empheq}
\usepackage{bm}
\usepackage{algorithm,algorithmic}
\usepackage{tkz-euclide}  %
\usepackage{multirow}
\usetikzlibrary{fadings}
\usepackage[symbol]{footmisc}

\usetkzobj{all}

\graphicspath{{figs/}}

\usepackage[super]{nth}
\definecolor{xkcd:olive}{HTML}{6E750E}
\definecolor{xkcd:tan}{HTML}{D1B26F}

\begin{document} 

\twocolumn[
\aistatstitle{Negative Momentum for Improved Game Dynamics}
\aistatsauthor{
   Gauthier Gidel$^*$ \quad 
   Reyhane Askari Hemmat$^*$ \quad 
   Mohammad Pezeshki \\
   \textbf{R\'emi Le Priol \quad 
   Gabriel Huang \quad 
   Simon Lacoste-Julien$^{\dag,\ddagger}$ \quad 
   Ioannis Mitliagkas$^\dag$}
  }

\aistatsaddress{  
   Mila \& DIRO, Universit\'e de Montr\'eal \qquad 
   $^\dag$Canada CIFAR AI chair \qquad $^\ddag$CIFAR fellow
   }
]
\begin{abstract}
Games generalize the single-objective optimization paradigm by introducing different objective functions for different players.
Differentiable games often proceed by simultaneous or alternating gradient updates.
In machine learning, games are gaining new importance through formulations like generative adversarial networks (GANs) and actor-critic systems.
However, compared to single-objective optimization, game dynamics is more complex and less understood.
In this paper, we analyze gradient-based methods with momentum on simple games.
We prove that alternating updates are more stable than simultaneous updates.
Next, we show both theoretically and empirically that alternating gradient updates with a negative momentum term achieves convergence in a difficult toy adversarial problem, but also on the notoriously difficult to train saturating GANs.
\end{abstract}

\section{INTRODUCTION}

Recent advances in machine learning are largely driven by the success of gradient-based optimization methods for the training process.
A common learning paradigm is empirical risk minimization, where a (potentially non-convex) objective, that depends on the data, is minimized.
However, some recently introduced approaches require the joint minimization of several objectives.
For example, actor-critic methods can be written as a bi-level optimization problem~\citep{pfau2016connecting} and generative adversarial networks (GANs)~\citep{goodfellow2014generative} use a two-player game formulation.

Games generalize the standard optimization framework by introducing different objective functions for different optimizing agents, known as {\em players}.
We are commonly interested in finding a local {\em Nash equilibrium}: a set of parameters from which no player can (locally and unilaterally) improve its objective function.
Games with differentiable objectives often proceed by simultaneous or alternating gradient steps on the players' objectives.
Even though the dynamics of gradient based methods is well understood for minimization problems, new issues appear in multi-player games. For instance, some stable stationary points of the dynamics may not be (local) Nash equilibria~\citep{adolphs2018local,daskalakis2018limit}.

\begin{figure*}[t]
    \centering
    \begin{minipage}[][][b]{0.55\textwidth}
        \centering
        \begin{tabular}{lcccc}
          \toprule 
           Method & $\beta$ &  Bounded & Converges & Bound on $\Delta_t$ \\
          \midrule
          \multirow{3}{*}{
          \hspace{-4mm}
          \begin{tabular}{l}
               \textcolor{xkcd:tan}{Simult.} \\[1mm]
               Thm. \ref{thm:bilin_sim} 
          \end{tabular}
          } 
          & {\red >0} & {\red \xmark} & {\red \xmark} & \!\!\!\! $\Omega\left( (1+\eta^2 \sigma^2_{\max}(A))^t\right)$\\
          & 0 & {\red \xmark} & {\red \xmark} & $\Omega\left( (1+\eta^2 \sigma^2_{\max}(A))^t\right)$ \\
          & {\color{SkyBlue} <0} & {\red \xmark} & {\red \xmark} & $\Omega\left( (1+\nicefrac{\eta^2 \sigma^2_{\max}(A)}{17})^{t}\right)$\\
          \midrule
          \multirow{3}{*}{
          \hspace{-4mm}
          \begin{tabular}{l}
               \textcolor{xkcd:olive}{Altern.}\\[1mm]
               Thm. \ref{thm:bilin_alt} 
          \end{tabular}
          } 
          & {\red >0} & {\red \xmark} & {\red \xmark} & \!\!\!\!
          \textbf{Conjecture:} $\Omega\left((1+\beta^2)^t\right)$ \\ 
          & 0 & {\green \cmark} & {\red \xmark} & $\Theta \left( \Delta_0 \right) $\\
          & {\color{SkyBlue}<0} & {\green \cmark} & {\green \cmark} & $\mathcal O \left( \Delta_0 ( 1 -  \nicefrac{\eta^2\sigma^2_{\min}(A)}{16})^t  \right) $\\
          \bottomrule
        \end{tabular}
        \hfill
    \end{minipage}
    \hfill
    \begin{minipage}[][][b]{0.32\textwidth}
        \vspace{2mm}
        \hfill 
        \includegraphics[width=\textwidth]{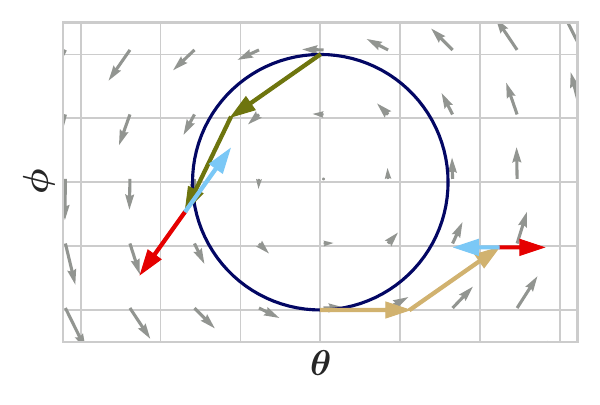}
        \vspace{-2mm}
    \end{minipage}
    \vspace{-4mm}
    \caption{
    \small \textbf{Left:} 
    Effect of gradient methods on an unconstrained bilinear example:
    $ \min_{\vtheta} \max_{\vphi} \;  \vtheta^\top \mA \vphi  \,.$ The quantity $\Delta_t$ is the distance to the optimum (see formal definition in \S\ref{sec:bilinear_game}) and $\beta$ is the momentum value.
    \textbf{Right:} 
    Graphical intuition of the role of momentum in two steps of simultaneous updates ({\color{xkcd:tan}\textbf{tan}}) or alternated updates ({\color{xkcd:olive}\textbf{olive}}). 
    Positive momentum ({\color{red}\textbf{red}}) drives the iterates outwards whereas negative momentum ({\color{SkyBlue}\textbf{blue}}) pulls the iterates back towards the center, but it is only strong enough for alternated updates.
    }
    \label{fig:convergence results}
\end{figure*}

Motivated by a decreasing trend of momentum values in GAN literature (see Fig.~\ref{fig:mom_trend}), we study the effect of two particular algorithmic choices:
(i) the choice between simultaneous and alternating updates, and (ii) the choice of step-size and momentum value.
The idea behind our approach is that a momentum term combined with the alternating gradient method can be used to manipulate the natural oscillatory behavior of adversarial games.
We summarize our main contributions:
\begin{itemize}
\item We prove in \S\ref{sec:bilinear_game} that the alternating gradient method with negative momentum is the only setting within our study parameters (Fig.~\ref{fig:convergence results}) that converges on a bilinear smooth game. Using a zero or positive momentum value, or doing simultaneous updates in such games fails to converge.
\item We show in \S\ref{sec:negative_momentum} that, for general dynamics, when the eigenvalues of the Jacobian have a large imaginary part, negative momentum can improve the local convergence properties of the gradient method.
\item We confirm the benefits of negative momentum for training GANs with the notoriously ill-behaved saturating loss on both toy settings, and real datasets. 
\end{itemize}

\paragraph{Outline.}
\S\ref{sec:background} describes the fundamentals of the analytic setup that we use. 
\S\ref{sec:tuning} provides a formulation for the optimal step-size, and discusses the constraints and intuition behind it.
\S\ref{sec:negative} presents our theoretical results and guarantees on negative momentum.
\S\ref{sec:bilinear_game} studies the properties of alternating and simultaneous methods with negative momentum on a bilinear smooth game. \S\ref{sec:results} contains experimental results on toy and real datasets. Finally, in \S\ref{sec:related}, we review some of the existing work on smooth game optimization as well as GAN stability and convergence.

\tikzset{every picture/.style={line width=0.75pt}} %

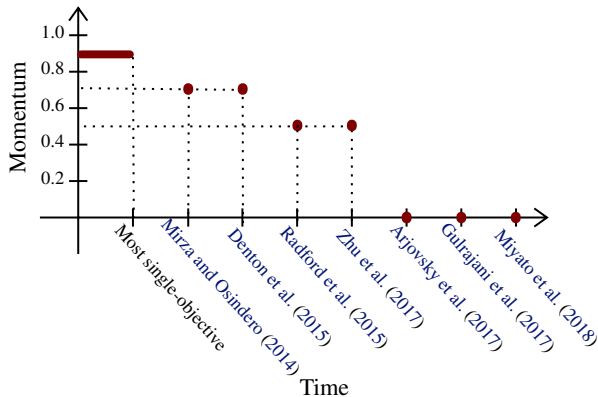
\begin{figure}
\centering
\begin{tikzpicture}[x=0.75pt,y=0.75pt,yscale=-1,xscale=1, scale=.92]
\draw    (330,212) -- (330,222) ;

\draw    (360,212) -- (360,222) ;

\draw    (390,212) -- (390,222) ;

\draw  (129,217) -- (407,217)(150,102) -- (150,237) (400,212) -- (407,217) -- (400,222) (145,109) -- (150,102) -- (155,109)  ;
\draw  [color={rgb, 255:red, 126; green, 0; blue, 0 }  ,draw opacity=1 ][line width=3] [line join = round][line cap = round] (151.53,127.5) .. controls (160.46,127.5) and (169.39,127.5) .. (178.32,127.5) ;
\draw  [color={rgb, 255:red, 126; green, 0; blue, 0 }  ,draw opacity=1 ][line width=3.75] [line join = round][line cap = round] (270.15,166) .. controls (270.15,166.33) and (270.15,166.67) .. (270.15,167) ;
\draw  [color={rgb, 255:red, 126; green, 0; blue, 0 }  ,draw opacity=1 ][line width=3.75] [line join = round][line cap = round] (330.04,216.45) .. controls (330.04,216.82) and (330.04,217.18) .. (330.04,217.55) ;
\draw  [color={rgb, 255:red, 126; green, 0; blue, 0 }  ,draw opacity=1 ][line width=3.75] [line join = round][line cap = round] (360,216.55) .. controls (360,216.85) and (360,217.15) .. (360,217.45) ;
\draw  [color={rgb, 255:red, 126; green, 0; blue, 0 }  ,draw opacity=1 ][line width=3.75] [line join = round][line cap = round] (389.88,216.55) .. controls (389.88,216.85) and (389.88,217.15) .. (389.88,217.45) ;
\draw  [dash pattern={on 0.84pt off 2.51pt}]  (180,129) -- (180.46,217) ;

\draw  [dash pattern={on 0.84pt off 2.51pt}]  (210.62,146) -- (210.31,219) ;

\draw  [dash pattern={on 0.84pt off 2.51pt}]  (240.15,146) -- (240.31,220) ;

\draw  [dash pattern={on 0.84pt off 2.51pt}]  (270.15,169) -- (270.15,221) ;

\draw  [dash pattern={on 0.84pt off 2.51pt}]  (300,169) -- (300,222) ;

\draw  [dash pattern={on 0.84pt off 2.51pt}]  (240.15,147) -- (149,146) ;

\draw  [dash pattern={on 0.84pt off 2.51pt}]  (300,167) -- (149.77,167) ;

\draw    (145,117) -- (155,117) ;

\draw    (145,137) -- (155,137) ;

\draw    (145,157) -- (155,157) ;

\draw    (145,177) -- (155,177) ;

\draw    (145,197) -- (155,197) ;

\draw    (180,212) -- (180,222) ;

\draw    (210.31,212) -- (210.31,222) ;

\draw    (240,212) -- (240,222) ;

\draw    (270,212) -- (270,222) ;

\draw    (300,212) -- (300,222) ;

\draw  [color={rgb, 255:red, 126; green, 0; blue, 0 }  ,draw opacity=1 ][line width=3.75] [line join = round][line cap = round] (240.15,146) .. controls (240.15,146.33) and (240.15,146.67) .. (240.15,147) ;
\draw  [color={rgb, 255:red, 126; green, 0; blue, 0 }  ,draw opacity=1 ][line width=3.75] [line join = round][line cap = round] (210.31,146) .. controls (210.31,146.33) and (210.31,146.67) .. (210.31,147) ;
\draw  [color={rgb, 255:red, 126; green, 0; blue, 0 }  ,draw opacity=1 ][line width=3.75] [line join = round][line cap = round] (300,166) .. controls (300,166.33) and (300,166.67) .. (300,167) ;

\draw (200.35,255.99) node [rotate=-49] [align=left] {{\fontfamily{ptm}\selectfont {\scriptsize Most single-objective}}};
\draw (233.32,260.99) node [rotate=-49] [align=left] {{\fontfamily{ptm}\selectfont {\scriptsize \citet{mirza2014conditional}}}};
\draw (260.29,255.99) node [rotate=-49] [align=left] {{\fontfamily{ptm}\selectfont {\scriptsize \citet{denton2015deep}}}};
\draw (290.17,255.99) node [rotate=-49] [align=left] {{\fontfamily{ptm}\selectfont {\scriptsize \citet{radford2015unsupervised}}}};
\draw (315.55,250.99) node [rotate=-49] [align=left] {{\fontfamily{ptm}\selectfont {\scriptsize \citet{zhu2017unpaired}}}};
\draw (351.61,255.99) node [rotate=-49] [align=left] {{\fontfamily{ptm}\selectfont {\scriptsize \citet{arjovsky2017wasserstein}}}};
\draw (379.49,255.99) node [rotate=-49] [align=left] {{\fontfamily{ptm}\selectfont {\scriptsize \citet{gulrajani2017improved}}}};
\draw (406.2,255.99) node [rotate=-49] [align=left] {{\fontfamily{ptm}\selectfont {\scriptsize \citet{miyato2018spectral}}}};
\draw (136,195) node  [align=left] {{\scriptsize {\fontfamily{ptm}\selectfont 0.2}}};
\draw (136,175) node  [align=left] {{\scriptsize {\fontfamily{ptm}\selectfont 0.4}}};
\draw (136,155) node  [align=left] {{\scriptsize {\fontfamily{ptm}\selectfont 0.6}}};
\draw (136,135) node  [align=left] {{\scriptsize {\fontfamily{ptm}\selectfont 0.8}}};
\draw (136,115) node  [align=left] {{\scriptsize {\fontfamily{ptm}\selectfont 1.0}}};
\draw (285,310) node  [align=left] {{\fontfamily{ptm}\selectfont {\footnotesize Time}}};
\draw (117,162) node [rotate=-270] [align=left] {{\fontfamily{ptm}\selectfont {\footnotesize Momentum}}};

\end{tikzpicture}
\caption{ \small Decreasing trend in the value of momentum used for training GANs across time.}
\vspace{-2mm}
\label{fig:mom_trend} 
\end{figure}

\section{BACKGROUND\label{sec:background}}

\paragraph{Notation}
In this paper, scalars are lower-case letters (e.g., $\lambda$), vectors are lower-case bold letters (e.g., $\vtheta$), matrices are upper-case bold letters (e.g., $\mA$) and operators are upper-case letters (e.g.,~$F$). The spectrum of a squared matrix $\mA$ is denoted by $\Sp(\mA)$, and its spectral radius is defined as $\rho(A)\defas \max \{|\lambda| \text{ for } \lambda\in \Sp(\mA) \}$. We respectively note $\sigma_{\min}(\mA)$ and $\sigma_{\max}(\mA)$ the smallest and the largest positive singular values of $\mA$. The identity matrix of $\R^{m \times m}$ is written $\mI_m$. We use $\Re$ and $\Im$ to respectively denote the real and imaginary part of a complex number.  $\mathcal O,$ $\Omega$ and $\Theta$ stand for the standard asymptotic notations. Finally, all the omitted proofs can be found in \S\ref{sec:proof_of_theorems}.

\paragraph{Game theory formulation of GANs}

Generative adversarial networks consist of a discriminator $D_{\vphi}$ and a generator $G_{\vtheta}$. 
In this game, the discriminator's objective is to tell real from generated examples.
The generator's goal is to produce examples that are sufficiently close to real examples to confuse the discriminator.

From a game theory point of view, GAN training is a differentiable two-player game: the discriminator $D_{\vphi}$ aims at minimizing its cost function $\LL_D$ and the generator $G_{\vtheta}$ aims at minimizing its own cost function $\LL_G$. 
Using the same formulation as the one in~\citet{mescheder_numerics_2017} and~\citet{gidel2018variational}, the GAN objective has the following form,
\begin{equation} \label{eq:two_player_games}
  \left\{
  \begin{aligned}
  \vtheta^* \in \argmin_{\vtheta \in \Theta}\LL_G(\vtheta,\bm{\vphi}^*) \,\\
  \bm{\vphi}^* \in \argmin_{\vphi \in \Phi} \LL_D(\vtheta^*,\vphi) \,. \\[-1mm]
  \end{aligned}
  \right.  
\end{equation}
Given such a game setup, GAN training consists of finding a local Nash Equilibrium, which is a state $(\vphi^*, \vtheta^*)$ in which neither the discriminator nor the generator can improve their respective cost by a small change in their parameters. In order to analyze the dynamics of gradient-based methods near a Nash Equilibrium, we look at the \textit{gradient vector field},
\begin{equation}
\bm{v}(\vphi, \vtheta) :=
\begin{bmatrix}
    \nabla_{\vphi}  \LL_D(\vphi, \vtheta) &
    \nabla_{\vtheta}  \LL_G(\vphi, \vtheta)
\end{bmatrix}^\top,
\end{equation}
and its associated \textit{Jacobian} $\nabla \bm{v}(\vphi, \vtheta)$,
\begin{equation}  
\begin{bmatrix}
    \nabla^2_{\vphi}  \LL_D(\vphi, \vtheta) & \nabla_{\vphi} \nabla_{\vtheta} \LL_D(\vphi, \vtheta)\\
    \nabla_{\vphi} \nabla_{\vtheta}  \LL_G(\vphi, \vtheta)^T & \nabla^2_{\vtheta}  \LL_G(\vphi, \vtheta)
\end{bmatrix}.
\end{equation}
Games in which $\LL_G = -\LL_D$ are called \textit{zero-sum games} and~\eqref{eq:two_player_games} can be reformulated as a min-max problem. This is the case for the original \textit{min-max} GAN formulation, but not the case for the \emph{non-saturating loss}~\citep{goodfellow2014generative} which is commonly used in practice.

For a zero-sum game, we note $\LL_G = -\LL_D = \LL$.
When the matrices $\nabla^2_{\vphi}  \LL(\vphi, \vtheta)$ and $\nabla^2_{\vtheta}  \LL(\vphi, \vtheta)$ are zero, the Jacobian is anti-symmetric and has pure imaginary eigenvalues. 
We call games with pure imaginary eigenvalues \emph{purely adversarial games}.
This is the case in a simple bilinear game $\LL(\vphi,\vtheta) := \vphi^\top \mA \vtheta$. This game can be formulated as a GAN where the true distribution is a Dirac on 0, the generator is a Dirac on $\theta$ and the discriminator is linear. This setup was extensively studied in 2D by \citet{gidel2018variational}.

Conversely, when $\nabla_{\vphi} \nabla_{\vtheta} \LL(\vphi, \vtheta)$ is zero and the matrices $\nabla^2_{\vphi}  \LL(\vphi, \vtheta)$ and $-\nabla^2_{\vtheta}  \LL(\vphi, \vtheta)$ are symmetric and definite positive, the Jacobian is symmetric and has real positive eigenvalues.
We call games with real positive eigenvalues \emph{purely cooperative games}. 
This is the case, for example, when the objective function $\LL$ is separable such as $\LL(\vphi,\vtheta) = f(\vphi) - g(\vtheta)$ where $f$ and $g$ are two convex functions. 
Thus, the optimization can be reformulated as two separated minimization of $f$ and $g$ with respect to their respective parameters. 

These notions of \emph{adversarial} and \emph{cooperative} games can be related to the notions of \emph{potential} games~\citep{monderer1996potential} and \emph{Hamiltonian} games recently introduced by~\citet{balduzzi2018mechanics}: a game is a \emph{potential game} (resp. \emph{Hamiltonian game}) if its Jacobian is symmetric (resp. asymmetric). Our definition of \emph{cooperative game} is a bit more general than the definition of \emph{potential game} since some non-symmetric matrices may have positive eigenvalues.
Similarly, the notion of \emph{adversarial game} generalizes the \emph{Hamiltonian games} since some non-antisymmetric matrices may have pure imaginary eigenvalues, for instance, 
\begin{equation}
    \Sp \left(\begin{bmatrix}
        0 & -1 \\ 2 & 3
    \end{bmatrix}
    \right) = \{1,2\} \,, \; \;
    \Sp \left(\begin{bmatrix}
        -1 & 1 \\ -2 & 1
    \end{bmatrix}
    \right) = \{\pm i\} \,. \notag
\end{equation}

In this work, we are interested in games \emph{in between} purely adversarial games and purely cooperative ones, i.e., games which have eigenvalues with non-negative real part (cooperative component) and non-zero imaginary part (adversarial component). For $\mA \in \mathbb{R}^{d \times p}$, a simple class of such games is parametrized by $\alpha \in [0,1]$,
\begin{equation} \label{eq:convex_adversarial_cooperative_games}
    \min_{\vtheta \in \mathbb{R}^d} \max_{\vphi \in \mathbb{R}^p} \; \alpha  \|\vtheta\|_2^2 + (1- \alpha) \vtheta^\top \mA \vphi - \alpha \|\vphi\|_2^2 \, ,
\end{equation}

\paragraph{Simultaneous Gradient Method.}
Let us consider the dynamics of the simultaneous gradient method. It is defined as the repeated application of the operator,
\begin{equation}
\label{eq:update}
F_{\eta}(\vphi, \vtheta) \defas \begin{bmatrix}
\vphi & \vtheta 
\end{bmatrix}^\top \!\!- \eta \ \bm{v}(\vphi, \vtheta) \,,  \quad (\vphi, \vtheta) \in \R^m\,,
\end{equation}
where $\eta$ is the learning rate.
Now, for brevity we write the joint parameters $\vomega \defas (\vphi, \vtheta) \in \R^m$.
For $t\in \mathbb{N}$,
let $\vomega_t = (\vphi_t, \vtheta_t)$
be the $t^\text{th}$ point of the sequence computed by the gradient method,
\begin{equation}
   \vomega_t = \underbrace{F_\eta \circ \ldots \circ F_\eta}_t (\vomega_0) =  F_\eta^{(t)} (\vomega_0) \,.
\end{equation}
Then, if the gradient method converges, and its limit point $\vomega^* = (\vphi^*,\vtheta^*)$ is a \emph{fixed point} of $F_\eta$ such that $\nabla v(\vomega^*)$ is positive-definite, then $\vomega^*$ is a local Nash equilibrium.
Interestingly, some of the stable stationary points of gradient dynamics may not be Nash equilibrium~\citep{adolphs2018local}. In this work, we focus on the local convergence properties near the stationary points of gradient .
To the best of our knowledge, there is no first order method alleviating this issue. In the following, $\vomega^*$ is a stationary point of the gradient dynamics (i.e. a point such that $\bm{v}(\vomega^*)=0$).

\section{TUNING THE STEP-SIZE\label{sec:tuning}} %
\label{sec:step_size_choice}

Under certain conditions on a fixed point operator,
linear convergence is guaranteed in a neighborhood around a fixed point.
\newcommand{\thmContractiveOperator}{
If the spectral radius $\rho_{\max} \defas \rho(\nabla F_\eta(\vomega^*)) < 1$, then,
for $\vomega_0$ in a neighborhood of $\vomega^*$,
the distance of $\vomega_t$ to the stationary point $\vomega^*$ converges at a linear rate of $\mathcal{O}\big((\rho_{\max}+\epsilon)^t\big) \,,\, \forall \epsilon >0$.
}
\begin{theorem}[Prop.~4.4.1 \citet{bertsekas1999nonlinear}]
\label{thm:contractive_operator}
\thmContractiveOperator
\end{theorem}

From the definition in \eqref{eq:update}, we have:
\begin{align}
  \nabla F_\eta (\vomega^*)  &= \mI_m - \eta \nabla \bm{v}(\vomega^*) \,, \\
   \text{and}\;\; \; \Sp(\nabla F_\eta (\vomega^*) )
  &= \left\{ 1 - \eta \lambda | \lambda \in \Sp(\nabla \bm{v}(\vomega^*)) \right\} \, . \notag
\end{align}
If the eigenvalues of $\nabla \bm{v}(\vomega^*)$ all have a positive real-part,
then for small enough $\eta$, the eigenvalues of $\nabla F_\eta(\vomega^*)$ are inside a convergence circle of radius $\rho_{\max} < 1$,
as illustrated in Fig.~\ref{fig:eigs}.
Thm.~\ref{thm:contractive_operator} guarantees the existence of an optimal step-size $\eta_{best}$ which yields a non-trivial convergence rate $\rho_{\max}<1$. 
Thm.~\ref{thm:best_step_size} gives analytic bounds on the optimal step-size $\eta_{best}$, and lower-bounds the best convergence rate $\rho_{\max}(\eta_{best})$ we can expect.

\newcommand{\thmBestStepSize}{ If the eigenvalues of $\nabla \bm{v}(\vomega^*)$ all have a positive real-part, then, the best step-size $\eta_{best}$, which minimizes the spectral radius $\rho_{\max}(\eta)$ of $\nabla F_\eta(\vphi^*, \vtheta^*)$, is the solution of a (convex) quadratic by parts problem, and satisfies,
\begin{align}
   &\max_{1 \leq k \leq m} \sin(\psi_k)^2
   \leq \rho_{\max}(\eta_{best})^2 \leq 1 - \Re(1/\lambda_1) \delta  \, ,\\
   &\text{with} \quad \delta := \min_{1\leq k\leq m}  |\lambda_k|^2(2 \Re(1/\lambda_k) - \Re(1/\lambda_1))\\
   &\text{and} \qquad 
   \Re(1/\lambda_1) \leq \eta_{best} \leq 2\Re(1/\lambda_1) 
 \end{align} 
 where 
 $(\lambda_k= r_k e^{i \psi_k})_{1\le k\leq m} = \Sp(\nabla \bm{v}(\vphi^*,\vtheta^*))$
 are sorted such that 
 $0<\Re(1/\lambda_1) \leq \cdots \leq \Re (1/\lambda_m)$.
 Particularly, when $\eta_{best} = \Re(1/\lambda_1) $
 we are in the case of the top plot of  Fig.\ref{fig:eigs} and 
 $\rho_{\max}(\eta_{best})^2 = \sin(\psi_1)^2 \; .$
 }
\begin{theorem}\label{thm:best_step_size}
   If the eigenvalues of $\nabla \bm{v}(\vomega^*)$ all have a positive real-part, then, the best step-size $\eta_{best}$, which minimizes the spectral radius $\rho_{\max}(\eta)$ of $\nabla F_\eta(\vphi^*, \vtheta^*)$, is the solution of a (convex) quadratic by parts problem, and satisfies,
\begin{align}
   &\max_{1 \leq k \leq m} \sin(\psi_k)^2
   \leq \rho_{\max}(\eta_{best})^2 \leq 1 - \Re(1/\lambda_1) \delta  \, , \label{eq:thm_best_step_size_rate}\\
   &\text{with} \quad \delta := \min_{1\leq k\leq m}  |\lambda_k|^2(2 \Re(1/\lambda_k) - \Re(1/\lambda_1)) \label{eq:def_delta} \\
   &\text{and} \qquad 
   \Re(1/\lambda_1) \leq \eta_{best} \leq 2\Re(1/\lambda_1)
   \label{eq:optim_step_size}
 \end{align} 
 where 
 $(\lambda_k= r_k e^{i \psi_k})_{1\le k\leq m} = \Sp(\nabla \bm{v}(\vphi^*,\vtheta^*))$
 are sorted such that 
 $0<\Re(1/\lambda_1) \leq \cdots \leq \Re (1/\lambda_m)$.
 Particularly, when $\eta_{best} = \Re(1/\lambda_1) $
 we are in the case of the top plot of  Fig.\ref{fig:eigs} and 
 $\rho_{\max}(\eta_{best})^2 = \sin(\psi_1)^2 \; .$
\end{theorem}
When $\nabla \bm{v}$ is positive-definite, the best $\eta_{best}$ is attained either because of one or several limiting eigenvalues. 
We illustrate and interpret these two cases in Fig.~\ref{fig:eigs}. 
In multivariate convex optimization, the optimal step-size depends on the extreme eigenvalues and their ratio, the {\em condition number}.
Unfortunately, the notion of the condition number does not trivially extend to games, but Thm.~\ref{thm:best_step_size} seems to indicate that the real part of the inverse of the eigenvalues play an important role in the dynamics of smooth games. We think that a notion of condition number might be meaningful for such games and we propose an illustrative example to discuss this point in \S\ref{sec:kappa_alpha}. Note that when the eigenvalues are pure positive real numbers belonging to $[\mu,L]$, \eqref{eq:thm_best_step_size_rate} provides the standard bound $\rho_{\max} \leq 1 - \mu/L$ obtained with a step-size $\eta = 1/L$ (see \S\ref{sub:thm:best_step_size} for details). 

\usetikzlibrary{calc,angles}

\begin{figure}[t]   
\vspace{-2mm}

\centering
\begin{subfigure}{}
\scalebox{.9}{
\begin{tikzpicture}[scale=1.6]
\def\pointsize{0.05}
\def\mucircle{0.6}

\draw[->,thick] (-1.5,0) -- (1.5,0) node[above] {\small $\mathbf{Re}$};
\draw[->,thick] (0,-1) -- (0,1.5) node[right] {\small$\mathbf{Im}$};

\draw[black,thick] (0,0) circle (1);
\draw[red!100!black,dashed,thick] (0,0) circle (\mucircle);

\def\arrowlength{0.43};

\draw ($(1,0)$) node[below right]{$1+0i$};
\fill ($(1,0)$) circle (\pointsize);

\def\maxx{-0.55}
\def\maxy{1.15}
\coordinate (a) at (0, 0) {};
\draw[-,thick] (1,0) -- (-1,0);

\foreach \x/\y/\name in {\maxx/\maxy/{1}, -1.2/0.6/{2}} {
\draw[dashed,gray] (\x,\y) coordinate (c) -- (1, 0) coordinate (I);

\fill[orange!80!black] (\x,\y) circle (\pointsize);
\draw (\x,\y) node[left]{$1-\lambda_\name$};

\fill[green!80!black] ($(1,0)!\arrowlength!(\x,\y)$) circle (\pointsize);

\draw ($(1,0)!\arrowlength!(\x,\y)$) node[left]{\tiny $1-\eta\lambda_\name$};
\draw   pic[draw=black, <-, angle eccentricity=1, angle radius=.4*\name cm]{angle = c--I--a};
}
\foreach \x/\y/\name in { -1.3/-0.6/{3}} {
\draw[dashed,gray] (\x,\y) coordinate (c) -- (1, 0) coordinate (I);

\fill[orange!80!black] (\x,\y) circle (\pointsize);
\draw (\x,\y) node[left]{$1-\lambda_\name$};

\fill[green!80!black] ($(1,0)!\arrowlength!(\x,\y)$) circle (\pointsize);

\draw ($(1,0)!\arrowlength!(\x,\y)$) node[left]{\tiny $1-\eta\lambda_\name$};
\draw   pic[draw=black, ->, angle eccentricity=1, angle radius=.4*\name cm]{angle = a--I--c};

}

\node at (.66,0.15)  {\tiny$\psi_1$};
\node at (.37,.09) {\tiny$\psi_2$};
\node at (.12,-.1) {\tiny$\psi_3$};
        
        \coordinate (inter) at (0.33, 0.49) {};
        \coordinate (O) at (0, 0) {};
        
\tkzMarkRightAngle[draw=black,size=.1](O,inter,I);
\draw[black](0,0) -- (0.33,0.49);

\end{tikzpicture}
}
\end{subfigure}
\begin{subfigure}{}
\scalebox{.9}{
\begin{tikzpicture}[scale=1.6]
\def\pointsize{0.05}
\def\mucircle{0.9}

\draw[->,thick] (-1.5,0) -- (1.5,0) node[above] {\small $\mathbf{Re}$};
\draw[->,thick] (0,-1) -- (0,1.5) node[right] {\small$\mathbf{Im}$};

\draw[black,thick] (0,0) circle (1);
\draw[red!100!black,dashed,thick] (0,0) circle (\mucircle);

\def\arrowlength{0.44};

\draw ($(1,0)$) node[below right]{$1+0i$};
\fill ($(1,0)$) circle (\pointsize);

\foreach \x/\y/\name in {-1/2/{\lambda_1}, -1.2/0.6/{\lambda_2}, 0.7/-0.3/{\lambda_3}} {
\draw[dashed,gray] (\x,\y) -- (1, 0);

\fill[orange!80!black] (\x,\y) circle (\pointsize);
\draw (\x,\y) node[left]{$1-\name$};

\fill[green!80!black] ($(1,0)!\arrowlength!(\x,\y)$) circle (\pointsize);

\draw ($(1,0)!\arrowlength!(\x,\y)$) node[left]{\tiny $1-\eta\name$};

}

\end{tikzpicture}
}
\end{subfigure}
\caption{ \small
Eigenvalues $\lambda_i$ of the Jacobian $\nabla \bm{v}(\bm{\phi^*},\bm{\theta^*})$ and their trajectories $1-\eta\lambda_i$ for growing step-sizes.
The unit circle is drawn in \textbf{black}, and the {\color{red} \textbf{red}} dashed circle  has radius equal to the largest eigenvalue $\mu_{\max}$, which is directly related to the convergence rate. Therefore, smaller red circles mean better convergence rates. \textbf{Top:} The red circle is limited by the tangent trajectory line $1-\eta\lambda_1$, which means the best convergence rate is limited only by the eigenvalue which will pass furthest from the origin as $\eta$ grows, i.e., $\lambda_i = \arg\min \Re(1/\lambda_i)$. \textbf{Bottom:} The red circle is cut (not tangent) by the trajectories at points $1-\eta\lambda_1$ and $1-\eta\lambda_3$. The $\eta$ is optimal because any increase in $\eta$ will push the eigenvalue $\lambda_1$ out of the red circle, while any decrease in step-size will retract the eigenvalue $\lambda_3$ out of the red circle, which will lower the convergence rate in any case. \emph{Figure inspired by~\citet{mescheder_numerics_2017}}.}
\vspace{-4mm}
\label{fig:eigs} 
\end{figure}
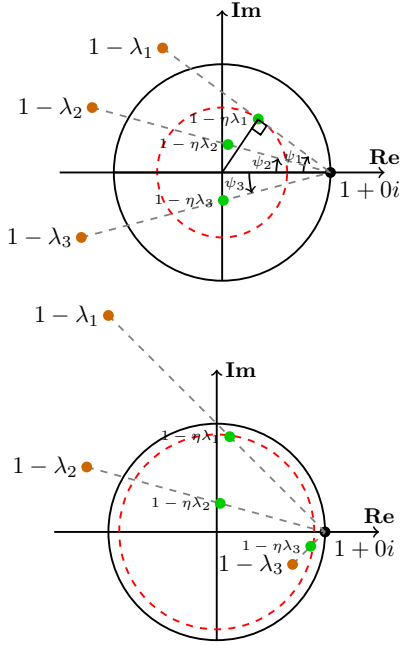

Note that, in~\eqref{eq:def_delta}, we have  $\delta > 0$ because $(\lambda_k)$ are sorted such that, $\Re(1/\lambda_k)\geq \Re(1/\lambda_1) \,,\, \forall 1\leq k\leq m$.
In~\eqref{eq:thm_best_step_size_rate}, we can see that if the Jacobian of $\bm{v}$ has an almost purely imaginary eigenvalue $r_j e^{\psi_j}$ then $\sin(\psi_j)$ is close to $1$ and thus, the convergence rate of the gradient method may be arbitrarily close to 1.
 \citet{zhang2017yellowfin} provide an analysis of the momentum method for quadratics, showing that momentum can actually help to better condition the model. 
 One interesting point from their work is that the best conditioning is achieved when the added momentum makes the Jacobian eigenvalues turn from positive reals into complex conjugate pairs.
 Our goal is to use momentum to wrangle game dynamics into convergence manipulating the eigenvalues of the Jacobian.

\section{NEGATIVE MOMENTUM\label{sec:negative}}
\label{sec:negative_momentum}
As shown in~\eqref{eq:thm_best_step_size_rate}, the presence of eigenvalues with large imaginary parts can restrict us to small step-sizes and lead to slow convergence rates.
In order to improve convergence, we add a \emph{negative} momentum term into the update rule.
Informally, one can think of negative momentum as friction that can damp oscillations.
The new momentum term leads to a modification of the \emph{parameter update operator} $F_\eta(\vomega)$ of~\eqref{eq:update}.
We use a similar state augmentation as~\cite{zhang2017yellowfin} and~\citet{daskalakis2018limit} to form a compound state $(\vomega_t,\vomega_{t-1}) := (\vphi_t,\vtheta_t,\vphi_{t-1}, \vtheta_{t-1}) \in \R^{2m}$. 
The update rule~\eqref{eq:update} turns into the following,
\begin{align}
  & F_{\eta,\beta}(\vomega_t,\vomega_{t-1}) = \left( \vomega_{t+1},\vomega_t \right) \\
  \text{where} \quad   
  & \vomega_{t+1} 
  \defas \vomega_t - \eta \vv(\vomega_t) + \beta (\vomega_t-\vomega_{t-1}) \,,
\end{align}
in which $\beta \in \mathbb{R}$ is the momentum parameter. Therefore, the Jacobian of $F_{\eta,\beta}$ has the following form,
\begin{equation}
  \begin{bmatrix}
    \bm{I}_n & \bm{0}_n \\ \bm{I}_n & \bm{0}_n
  \end{bmatrix}
  - \eta \begin{bmatrix}
    \nabla \bm{v}(\vomega_t) & \bm{0}_n \\ \bm{0}_n & \bm{0}_n
    \end{bmatrix}    
    + \beta
    \begin{bmatrix}
    \bm{I}_n & - \bm{I}_n \\ \bm{0}_n & \bm{0}_n
    \end{bmatrix}
\end{equation}
Note that for $\beta=0$, we recover the gradient method.

\begin{figure}
\centering
\includegraphics[width = .28\textwidth]{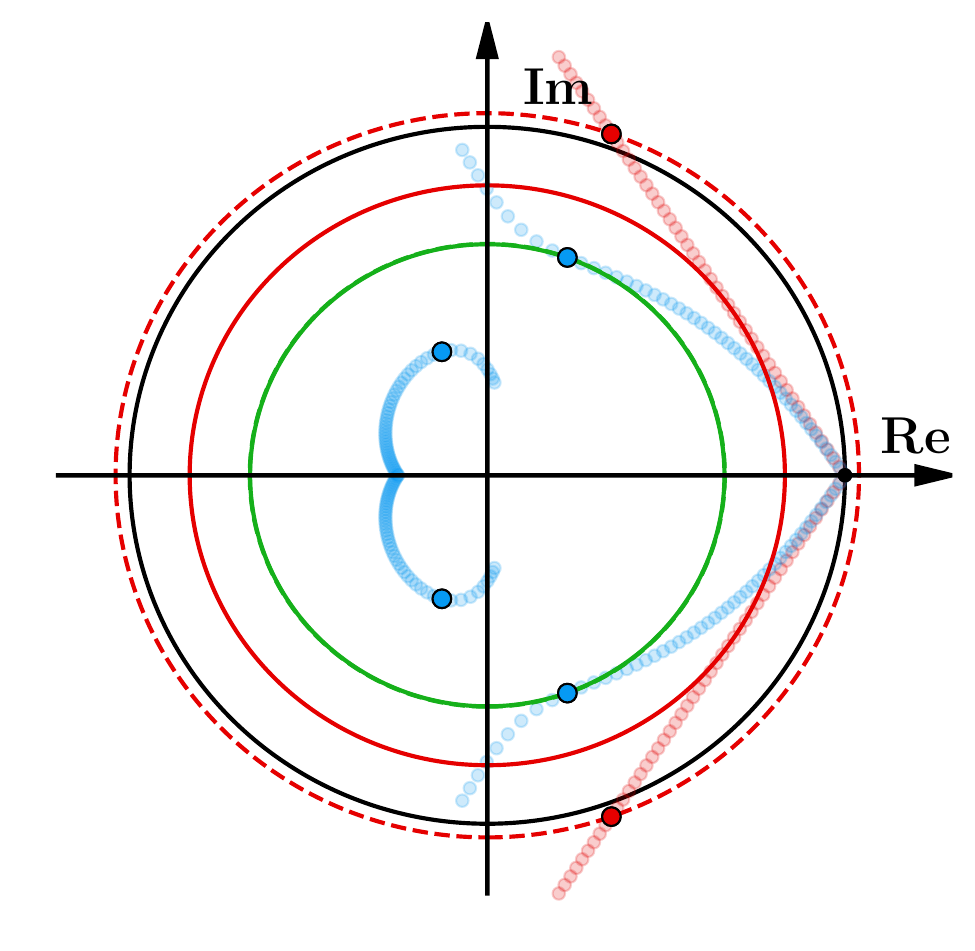}
\caption{\small
Transformation of the eigenvalues by a negative momentum method for a game introduced in \eqref{eq:convex_adversarial_cooperative_games} with $d=p=1, A=1, \alpha=.4, \eta=1.55,\beta=-.25$.
Convergence circles for gradient method are in {\color{red}\textbf{red}}, negative momentum in {\color{bettergreen}\textbf{green}}, and unit circle in \textbf{black}. \underline{Solid} convergence circles are optimized over all step-sizes, while \underline{d}a\underline{s}h\underline{e}d circles are at a given step-size $\eta$. For a fixed $\eta$, original eigenvalues are in {\color{red}\textbf{red}} and negative momentum eigenvalues are in {\color{blue}\textbf{blue}}. Their trajectories as $\eta$ sweeps in $[0,2]$ are in light colors.
Negative momentum helps as the new convergence circle (green) is smaller, due to shifting the original eigenvalues (red dots) towards the origin (right blue dots), while the eigenvalues due to state augmentation (left blue dots) have smaller magnitude and do not influence the convergence rate. Negative momentum allows faster convergence (green circle inside the solid red circle) for a broad range of step-sizes. 
\label{fig:effect}
}
\end{figure}
In some situations, if $\beta<0$ is adjusted properly, negative momentum can improve the convergence rate to a local stationary point by pushing the eigenvalues of its Jacobian towards the origin. In the following theorem, we provide an explicit equation for the eigenvalues of the Jacobian of $F_{\eta,\beta}$. 

\newcommand{\thmEigenMomentum}{
The eigenvalues of $\nabla F_{\eta,\beta}(\vomega^*)$ are
\begin{equation}
   \mu_{\pm}(\beta,\eta,\lambda):= (1 -\eta \lambda + \beta)\frac{1 \pm \Delta^{\frac{1}{2}}}{2},
\end{equation}
where $ \Delta := 1 - \frac{4 \beta}{(1 - \eta \lambda + \beta)^2}  \, , \; \lambda \in \Sp(\nabla\bm{v}(\vomega^*))$ and $\Delta^{\frac{1}{2}}$ is the \emph{complex} square root of $\Delta$ with positive real part\footnote{ If $\Delta$ is a negative real number we set $\Delta^{\frac{1}{2}}:= i \sqrt{-\Delta}$}.  Moreover we have the following Taylor approximation,
\begin{align}
   \mu_{+}(\beta,\eta,\lambda) &= 1 - \eta \lambda - \beta \frac{\eta \lambda}{1 - \eta \lambda} + O(\beta^2) \, , \\
   \mu_{-}(\beta,\eta,\lambda) &= \frac{\beta}{1-\eta\lambda} + O(\beta^2) \,.
\end{align}
}
\begin{theorem}
\label{thm:eigen_F_eta_beta}
\thmEigenMomentum
\end{theorem}

When $\beta$ is small enough, $\Delta$ is a complex number close to $1$. 
Consequently, $\mu_{+}$ is close to the original eigenvalue for gradient dynamics $ 1 -\eta \lambda$, 
and $\mu_{-}$, the eigenvalue introduced by the state augmentation, is close to 0. 
We formalize this intuition by providing the first order approximation of both eigenvalues.

In Fig.~\ref{fig:effect}, we illustrate the effects of negative momentum on
a  game described in~\eqref{eq:convex_adversarial_cooperative_games}. Negative momentum shifts the original eigenvalues (trajectories in light red) by pushing them to the left towards the origin (trajectories in light blue). 

Since our goal is to minimize the largest magnitude of the eigenvalues of $F_{\eta,\beta}$ which are computed in Thm.~\ref{thm:eigen_F_eta_beta}, we want to understand the effect of $\beta$ on these eigenvalues with potential large magnitude. 
Let $\lambda \in \Sp(\nabla\bm{v}(\vomega^*))$, we define the (squared) magnitude $\rho_{\lambda,\eta}(\beta)$ that we want to optimize,
\begin{equation}
  \label{eq:def_magnitude}
  \rho_{\lambda,\eta}(\beta) := \max\left\{ 
  |\mu_{+}(\beta,\eta,\lambda)|^2,|
  \mu_{-}(\beta,\eta,\lambda)|^2
  \right\}.
\end{equation}
We study the local behavior of $\rho_{\lambda,\eta}$ for small $\beta$. The following theorem shows that a well suited $\beta$ decreases $\rho_{\lambda,\eta}$, which corresponds to faster convergence.

\newcommand{\thmDerivativePositive}{
For any $\lambda \in \Sp(\nabla\bm{v}(\vomega^*))$ s.t. $\Re(\lambda)>0$,
\begin{equation}\notag
  \rho_{\lambda,\eta}'(0) > 0 \Leftrightarrow
 \eta \in I(\lambda) \defas \left(\tfrac{|\lambda|-|\Im(\lambda)|}{|\lambda| \Re(\lambda)}, \tfrac{|\lambda|+ |\Im(\lambda)|}{|\lambda|\Re(\lambda)}\right) \,.
\end{equation}
Particularly, we have $\rho_{\lambda,\Re(1/\lambda)}'(0) = 2 \Re(\lambda) \Re(1/\lambda) >0$ and 
$|\text{Arg}(\lambda)| \geq \frac{\pi}{4} \Rightarrow \left( \Re({1}/{\lambda}) ,2 \Re({1}/{\lambda})\right) \subset I(\lambda) $.
}

\begin{theorem}\label{thm:derivative_positive}
\thmDerivativePositive
\end{theorem}
As we have seen previously in Fig.~\ref{fig:eigs} and Thm.~\ref{thm:best_step_size}, there are only few eigenvalues which slow down the convergence.
Thm.~\ref{thm:derivative_positive} is a local result showing that a small  negative momentum can improve the magnitude of the limiting eigenvalues in the following cases: 
when there is only one limiting eigenvalue $\lambda_1$ (since  in that case the optimal step-size is $\eta_{best}=\Re(1/\lambda_1) \in I(\lambda_1)$) 
or when there are several limiting eigenvalues $\lambda_1,\ldots,\lambda_k$ and the intersection $I(\lambda_1) \cap \ldots \cap I(\lambda_k)$ is not empty.
We point out that we do not provide any guarantees on whether this intersection is empty or not but note that if the absolute value of the argument of $\lambda_1$ is larger than $\pi/4$ then by~\eqref{eq:optim_step_size}, our theorem provides that the optimal step-size $\eta_{best}$ belongs to $I(\lambda_1)$.

Since our result is local, it does not provide any guarantees on large negative values of $\beta$. 
Nevertheless, we numerically optimized~\eqref{eq:def_magnitude} with respect to $\beta$ and $\eta$ and found that 
for any non-imaginary fixed eigenvalue $\lambda$, 
the optimal momentum is negative 
and the associated optimal step-size is larger than $\hat \eta(\lambda)$.
Another interesting aspect of negative momentum is that it admits larger step-sizes (see Fig.~\ref{fig:effect} and~\ref{fig:lr_mom}).

For a game with purely imaginary eigenvalues, when $|\eta \lambda| \ll 1$, Thm.~\ref{thm:eigen_F_eta_beta} shows that $\mu_{+}(\beta,\eta,\lambda) \approx 1 - (1+\beta)\eta \lambda$. Therefore, at the first order, $\beta$ only has an impact on the imaginary part of $\mu_+$. Consequently $\mu_+$ cannot be pushed into the unit circle, and the convergence guarantees of Thm.~\ref{thm:contractive_operator} do not apply.
In other words, the analysis above provides convergence rates for games without any pure imaginary eigenvalues. 
It excludes the purely adversarial bilinear example ($\alpha = 0$ in Eq.~\ref{eq:convex_adversarial_cooperative_games})
that is discussed in the next section. 

\section{BILINEAR SMOOTH GAMES} %
\label{sec:bilinear_game}

In this section we analyze the dynamics of a purely adversarial game described by,
\begin{equation} \label{eq:adversarial_games}
    \min_{\vtheta \in \mathbb{R}^d} \max_{\vphi \in \mathbb{R}^p} \;  \vtheta^\top \mA \vphi + \vtheta^\top \vb + \vc^\top \vphi, \quad \mA \in \mathbb{R}^{d \times p} \,.
\end{equation}
The first order stationary condition for this game characterizes the solutions $(\vtheta^*,\vphi^*)$ as
\begin{equation}
    \mA \vphi^* = \vb \quad \text{and} \quad \mA^\top \vtheta^* = \vc \,.
\end{equation}
If $\vb$ (resp. $\vc$) does not belong to the column space of $\mA$ (resp. $\mA^\top$), the game \eqref{eq:adversarial_games} admits no equilibrium. 
In the following, we assume that an equilibrium does exist for this game. Consequently, there exist $\vb'$ and $\vc'$ such that $\vb = \mA \vb'$ and $\vc = \mA^\top \vc'$. Using the translations $\vtheta \rightarrow \vtheta - \vc'$ and $\vphi \rightarrow \vphi - \vb'$, we can assume without loss of generality, that $p \geq d$, $\vb = \bm{0}$ and $\vc = \bm{0}$.
We provide upper and lower bounds on the squared distance from the known equilibrium,
\begin{equation}
    \Delta_{t} = \|\vtheta_t-\vtheta^*\|_2^2 + \|\vphi_t-\vphi^*\|_2^2
\end{equation}
where $(\vtheta^*,\vphi^*)$ is the projection of ($\vtheta_t,\vphi_t)$ onto the solution space.
We show in \S\ref{sec:lemmas_and_definitions}, Lem.~\ref{lemma:reduc_dim} that, for our methods of interest, this projection has a simple formulation that only depends on the initialization $(\vtheta_0,\vphi_0)$.

We aim to understand the difference between the dynamics of simultaneous steps and alternating steps.
Practitioners have been widely using the latter instead of the former when optimizing GANs despite the rich optimization literature on simultaneous methods.
\subsection{Simultaneous gradient descent}
We define this class of methods with momentum using the following formulas,
\begin{align} \label{eq:updates_sim}
  &\! \! \quad F^{\text{sim}}_{\eta,\beta}(\vtheta_t , \vphi_{t} , \vtheta_{t-1},\vphi_{t-1})
    \defas   ( \vtheta_{t+1} , \vphi_{t+1} , \vtheta_{t}, \vphi_{t} ) \\[2mm]
  &\text{where} \;  
    \left\{
    \begin{aligned}
        \vtheta_{t+1}
        &=  \vtheta_t - \eta_1 \mA \vphi_t + \beta_1 (\vtheta_t  -\vtheta_{t-1} )\\ \! 
        \vphi_{t+1} 
        &= \vphi_t + \eta_2 \mA^\top\vtheta_{t} + \beta_2(\vphi_t  -\vphi_{t-1} ) \,.  \! \!\! \!
    \end{aligned}\notag
      \right. 
\end{align}
In our simple setting, the operator $F^{\text{sim}}_{\eta,\beta}$ is linear. One way to study the asymptotic properties of the sequence $(\vtheta_t,\vphi_t)$ is to compute the eigenvalues of $\nabla 
F^{\text{sim}}_{\eta,\beta}$. The following proposition characterizes these eigenvalues.
\newcommand{\propEigsSim}{
The eigenvalues of $\nabla F^{\text{sim}}_{\eta,\beta}$ are the roots of the \nth{4} order polynomials:
\begin{equation}
     (x-1)^2(x-\beta_1)(x-\beta_2) + \eta_1 \eta_2\lambda x^2 \,, \, \lambda \in \Sp(\mA^\top \mA).
\end{equation}
}
\begin{proposition}\label{prop:eigs_bilinear_sim}
 The eigenvalues of $\nabla F^{\text{sim}}_{\eta,\beta}$ are the roots of the \nth{4} order polynomials:
\begin{equation}
     (x-1)^2(x-\beta_1)(x-\beta_2) + \eta_1 \eta_2\lambda x^2 , \, \lambda \in \Sp(\mA^\top \mA) \label{eq:poly_simultaneous}
\end{equation}
\end{proposition}

Interestingly, these roots only depend on the product $\eta_1 \eta_2$ meaning that any re-scaling $\eta_1 \rightarrow \gamma \eta_1 \, , \; \eta_2 \rightarrow \frac{1}{\gamma} \eta_2$ does not change the eigenvalues of $\nabla F^{\text{sim}}_{\eta,\beta}$ and consequently the asymptotic dynamics of the iterates $(\vtheta_t,\vphi_t)$. 
The magnitude of the eigenvalues described in~\eqref{eq:poly_simultaneous}, characterizes the asymptotic properties for the iterates of the simultaneous method~\eqref{eq:updates_sim}.
We report the maximum magnitude of these roots for a given $\lambda$ and for a grid of step-sizes and momentum values in Fig~\ref{fig:magnitudes}.
We observe that they are always larger than 1, which transcribes a diverging behavior.
The following theorem provides an analytical rate of divergence.

\begin{figure*}
\centering
\vspace{-2mm}
      \includegraphics[width=0.74\textwidth]{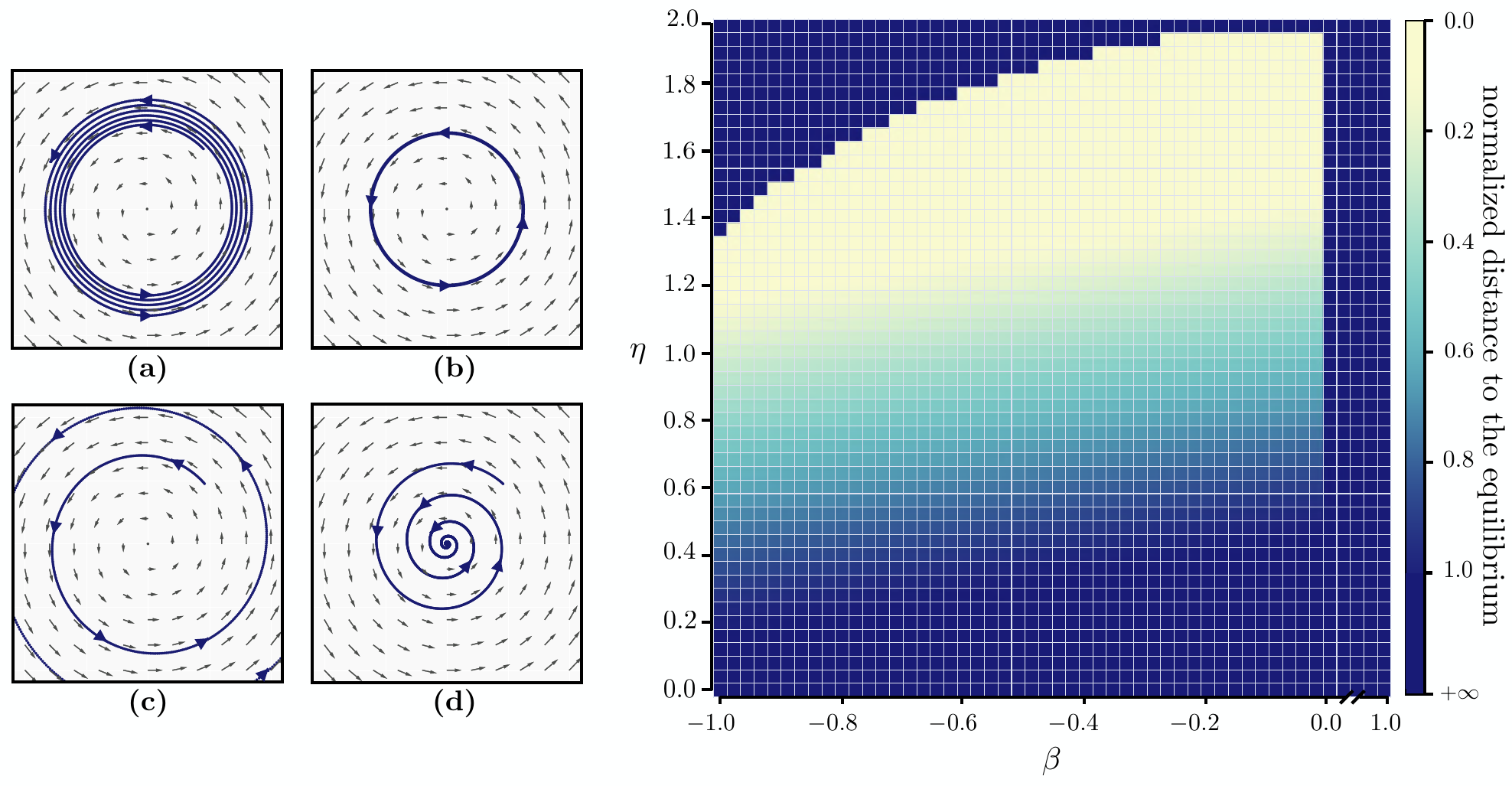}
      \vspace{-4mm}
  \caption{ 
  \small The effect of momentum in a simple min-max bilinear game where the equilibrium is at $(0, 0)$. \textbf{(left-a)} Simultaneous GD with no momentum \textbf{(left-b)} Alternating GD with no momentum. \textbf{(left-c)} Alternating GD with a momentum of $+0.1$. \textbf{(left-d)} Alternating GD with a momentum of $-0.1$. \textbf{(right)} A grid of experiments for alternating GD with different values of momentum ($\beta$) and step-sizes ($\eta$): While any positive momentum leads to divergence, small enough value of negative momentum allows for convergence with large step-sizes. The color in each cell indicates the normalized distance to the equilibrium after 500k iteration, such that $1.0$ corresponds to the initial condition and values larger (smaller) than $1.0$ correspond to divergence (convergence).}
\label{fig:lr_mom}
\vspace{-2mm}
\end{figure*}

\newcommand{\thmDivBilin}{
For any $\eta_1, \eta_2 \geq 0$ and $\beta_1=\beta_2 =\beta$,  the iterates of the simultaneous methods~\eqref{eq:updates_sim} diverge as,
\begin{equation}\notag
  \Delta_{t} \in\left\{ 
  \begin{aligned}
  &\Omega\big(\Delta_0(1 + \eta^2 \sigma^2_{\max}(A))^t\big) \;\ \text{if} \quad  \beta \geq 0\\
  & \Omega\big(\Delta_0(1 + \tfrac{\eta^2 \sigma^2_{\max}(A)}{17})^t\big) \;\; \text{if} \quad  -\frac{1}{16} \leq \beta <0 \,.
  \end{aligned}\right.
\end{equation}  
}
\begin{theorem}
\label{thm:bilin_sim}
    \thmDivBilin
\end{theorem}
This theorem states that the iterates of the simultaneous method~\eqref{eq:updates_sim} diverge geometrically for $\beta \geq -\tfrac{1}{16}$.
Interestingly, this geometric divergence implies that \textit{even} a uniform averaging of the iterates (standard in game optimization to ensure convergence~\citep{freund1999adaptive}) cannot alleviate this divergence.

\subsection{Alternating gradient descent}
Alternating gradient methods take advantage of the fact that the iterates $\vtheta_{t+1}$ and $\vphi_{t+1}$ are computed sequentially, to plug the value of $\vtheta_{t+1}$ (instead of $\vtheta_t$ for simultaneous update rule) into the update of $\vphi_{t+1}$,
\begin{align}\label{eq:updates_alt}
  \hspace*{-5cm} F^{\text{alt}}_{\eta,\beta} (\vtheta_t , \vphi_{t} , \vtheta_{t-1},\vphi_{t-1})
    \!\defas\!  ( \vtheta_{t+1} , \vphi_{t+1} , \vtheta_{t}, \vphi_{t} )\!\! \\[2mm]
 \notag
 \!\!\!\text{where} 
  \left\{\begin{aligned}\vtheta_{t+1}&=  \vtheta_t - \eta_1 \mA \vphi_t + \beta_1 (\vtheta_t  -\vtheta_{t-1} )\\ 
      \vphi_{t+1}  &= \vphi_t +  
      \eta_2 \mA^\top \vtheta_{t+1} + \beta_2 (\vphi_t  -\vphi_{t-1} ) \,. \!\!\!\! \!
      \end{aligned}
      \right. \!\!
\end{align}
This slight change between~\eqref{eq:updates_sim} and~\eqref{eq:updates_alt} significantly shifts the eigenvalues of the Jacobian. We first characterize them with the following proposition.

\newcommand{\propEigsAlt}{
The eigenvalues of $\nabla F^{\text{alt}}_{\eta,\beta}$ are the roots of the \nth{4} order polynomials:
\begin{equation}
 (x-1)^2(x-\beta_1)(x-\beta_2)+ \eta_1 \eta_2 \lambda x^3 \, , \,\lambda \in \Sp(\mA^\top \mA)
\end{equation}
}

\begin{proposition}
\label{prop:eigs_bilinear_alt}
The eigenvalues of $\nabla F^{\text{alt}}_{\eta,\beta}$ are the roots of the \nth{4} order polynomials:
\begin{equation}
 (x-1)^2(x-\beta_1)(x-\beta_2)+ \eta_1 \eta_2 \lambda x^3 \, , \,\lambda \in \Sp(\mA^\top \mA) \label{eq:poly_alternated}
\end{equation}
\end{proposition}
The same way as in~\eqref{eq:poly_simultaneous}, these roots only depend on the product $\eta_1\eta_2$. 
The only difference is that the monomial with coefficient $\eta_1\eta_2 \lambda$ is of degree 2 in~\eqref{eq:poly_simultaneous} and of degree 3 in~\eqref{eq:poly_alternated}. 
This difference is major since, for well chosen values of negative momentum, the eigenvalues described in Prop.~\ref{prop:eigs_bilinear_alt} lie in the unit disk (see \textbf{}Fig.~\ref{fig:magnitudes}). 
As a consequence, the iterates of the alternating method with no momentum are bounded and do converge if we add some well chosen negative momentum:
\newcommand{\thmConvBilin}{If we set $\eta \leq \frac{1}{\sigma_{\max}(A)}$, $\beta_1 = -\frac{1}{2}$ and $\beta_2 = 0$ then we have
  \begin{equation}
    \Delta_{t+1} \in O \left( \max\{\tfrac{1}{2},1- \tfrac{\eta^2\sigma^2_{\min}(A)}{16}\}^t \Delta_0 \right)
  \end{equation}
  If we set $\beta_1 = 0$ and $\beta_2 = 0$, then there exists $M>1$ such that for any $\eta_1,\eta_2 \geq 0$, $\Delta_t = \Theta(\Delta_0)$.
}
\begin{theorem}
\label{thm:bilin_alt}
  \thmConvBilin
\end{theorem}
Our results from this section, namely
Thm.~\ref{thm:bilin_sim} and Thm.~\ref{thm:bilin_alt}, are summarized in Fig.~\ref{fig:convergence results}, and demonstrate how alternating steps can improve the convergence properties of the gradient method for bilinear smooth games. Moreover, combining them with negative momentum can surprisingly lead to a linearly convergent method. The conjecture provided in Fig.~\ref{fig:convergence results} (divergence of the alternating method with positive momentum) is backed-up by the results provided in Fig.~\ref{fig:lr_mom} and \S\ref{sub:plotmag}.

\section{EXPERIMENTS AND DISCUSSION\label{sec:results}} %

\paragraph{Min-Max Bilinear Game\hspace{-2mm}}[Fig. \ref{fig:lr_mom}]\hspace{2mm}
In our first experiments, we showcase the effect of negative momentum in a bilinear min-max optimization setup~\eqref{eq:convex_adversarial_cooperative_games} where $\bm \phi, \bm \theta \in \R$ and $\mA = 1$. 
We compare the effect of positive and negative momentum in both cases of alternating and simultaneous gradient steps.

\begin{figure*}
\vspace{-2mm}
    \centering
    \includegraphics[width=.9\linewidth]{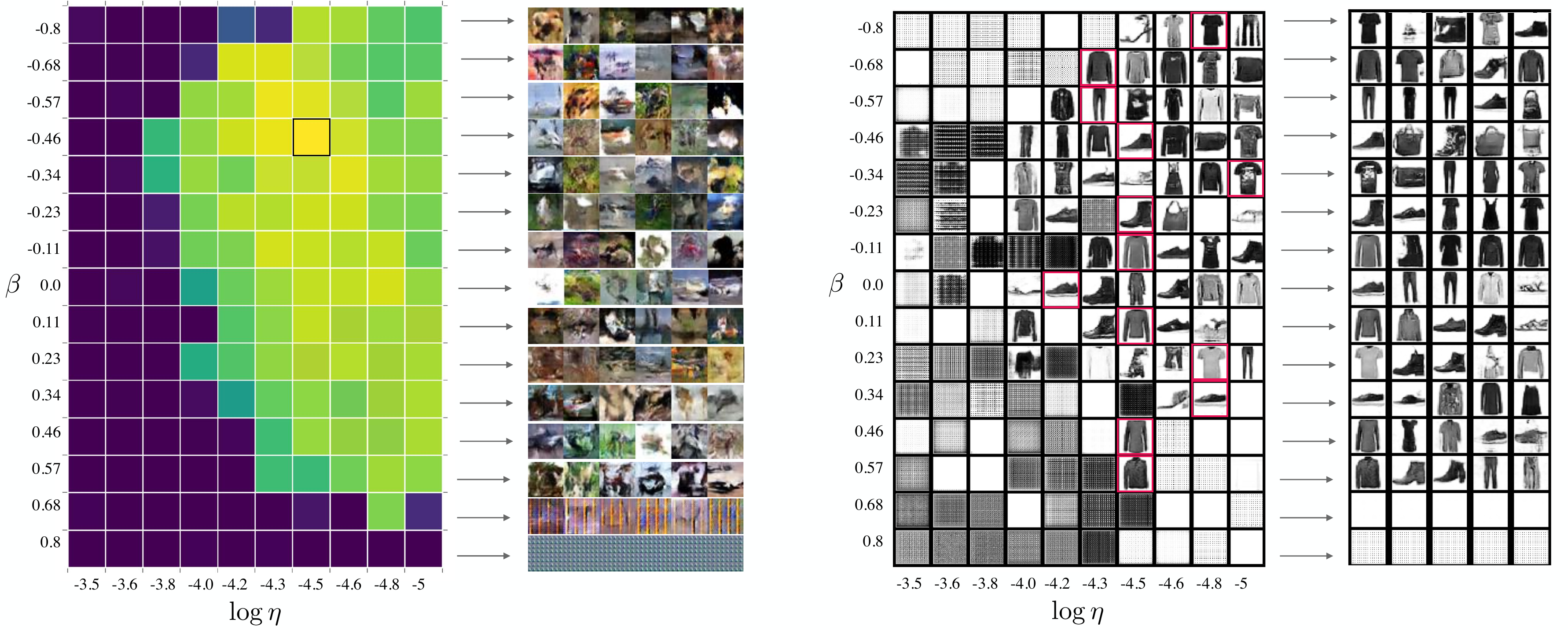}
    \vspace{-2mm}
    \caption{
    \small Comparison between negative and positive momentum on GANs with saturating loss on CIFAR-10 (left) and on Fashion MNIST (right) using a residual network. For each dataset, a grid of different values of momentum ($\beta$) and step-sizes ($\eta$) is provided which describes the discriminator's settings while a constant momentum of $0.5$ and step-size of $10^{-4}$ is used for the generator. Each cell in CIFAR-10 (or Fashion MNIST) grid contains a single configuration in which its color (or its content) indicates the inception score (or a single sample) of the model. For CIFAR-10 experiments, yellow is higher while blue is the lower inception score. Along each row, the best configuration is chosen and more samples from that configuration are presented on the right side of each grid.
    }
\label{fig:mnist_cifar}
\vspace{-3mm}
\end{figure*}

\paragraph{Fashion MNIST and CIFAR 10\hspace{-2mm}}[Fig.~\ref{fig:mnist_cifar}] \hspace{2mm}  In our third set of experiments, we use negative momentum in a GAN setup on CIFAR-10 \citep{krizhevsky2009learning} and Fashion-MNIST \citep{xiao2017fashion} with \emph{saturating loss} and alternating steps. We use residual networks for both the generator and the discriminator with no batch-normalization. Following the same architecture as \citet{gulrajani2017improved}, each residual block is made of two $3 \times 3$ convolution layers with \emph{ReLU} activation function. Up-sampling and down-sampling layers are respectively used in the generator and discriminator. We experiment with different values of momentum on the discriminator and a constant value of 0.5 for the momentum of the generator. We observe that using a negative value can generally result in samples with higher quality and inception scores. Intuitively, using negative momentum only on the discriminator slows down the learning process of the discriminator and allows for better flow of the gradient to the generator. Note that we provide an additional experiment on mixture of Gaussians in \S~\ref{sub:mog}.
\vspace{-2mm}
\section{RELATED WORK}
\label{sec:related}
\vspace{-2mm}

\paragraph{Optimization}

From an optimization point of view, a lot of work has been done in the context of understanding momentum and its variants \citep{polyak1964some,qian1999momentum,nesterov2013introductory,sutskever2013importance}. Some recent studies have emphasized the importance of momentum tuning in deep learning such as \citet{sutskever2013importance}, \cite{kingma2014adam}, and \citet{zhang2017yellowfin}, however, none of them consider using negative momentum. Among recent work, using robust control theory, \citet{lessard2016analysis} study optimization procedures and cover a variety of algorithms including momentum methods. Their analysis is global and they establish worst-case bounds for smooth and strongly-convex functions. 
\citet{mitliagkas2016asynchrony}~considered negative momentum in the context of asynchronous single-objective minimization. They show that asynchronous-parallel dynamics ‘bleed’ into optimization updates introducing momentum-like behavior into SGD. They argue that algorithmic momentum and asynchrony-induced momentum add up to create an effective ‘total momentum’ value. They conclude that to attain the optimal (positive) effective momentum in an asynchronous system, one would have to reduce algorithmic momentum to small or sometimes negative values. This differs from our work where we show that for games the optimal effective momentum may be negative.
\citet{ghadimi2015global} analyze momentum and provide global convergence properties for functions with Lipschitz-continuous gradients. 
However, all the results mentioned above are restricted to minimization problems. The purpose of our work is to try to understand how momentum influences game dynamics which is intrinsically different from minimization dynamics. 

Finally, similar proof techniques based on the study of the eigenvalues of a state-augmented operator have been recently used by~\citet{daskalakis2018limit} for the study of the optimistic gradient method (OGDA). However, even though OGDA and Polyak's momentum can be seen as a variant of the gradient method with an additional term, these additional terms are fundamentally different. In OGDA it is a difference between the two previous gradients, while in Polyak's method it is a difference between the two past iterates.  

\paragraph{GANs as games}
A lot of recent work has attempted to make GAN training easier with new optimization methods.
\citet{daskalakis2017training} extrapolate the next value of the gradient using previous history and
\citet{gidel2018variational} explore averaging and introduce a variant of the extra-gradient algorithm.
\citet{balduzzi2018mechanics} develop new methods to understand the dynamics of general games: they decompose second-order dynamics into two components using Helmholtz decomposition and use the fact that the optimization of Hamiltonian games is well understood. It differs from our work since we do not consider any decomposition of the Jacobian but focus on the manipulation of its eigenvalues. Recently, \citet{liang2018interaction} provide a unifying theory for smooth two-player games for non-asymptotic local convergence. They also provide theory for choosing the right step-size required for convergence. 

From another perspective,
\citet{odena2018generator} show that in a GAN setup, the average conditioning of the Jacobian of the generator becomes ill-conditioned during training. They propose Jacobian clamping to improve the inception score and Frechet Inception Distance.
\citet{mescheder_numerics_2017} provide discussion on how the eigenvalues of the Jacobian govern the local convergence properties of GANs. They argue that the presence of eigenvalues with zero real-part and large imaginary-part results in oscillatory behavior but do not provide results on the optimal step-size and on the impact of momentum. 
\citet{nagarajan_gradient_2017} also analyze the local stability of GANs as an approximated continuous dynamical system. They show that during training of a GAN, the eigenvalues of the Jacobian of the corresponding vector field are pushed away from one along the real axis. 

\section{CONCLUSION} %
In this paper, we show analytically and empirically that alternating updates with negative momentum is the only method within our study parameters (Fig.\ref{fig:convergence results}) that converges in bilinear smooth games. We study the effects of using negative values of momentum in a GAN setup both theoretically and experimentally. We show that, for a large class of adversarial games, negative momentum may improve the convergence rate of gradient-based methods by shifting the eigenvalues of the Jacobian appropriately into a smaller convergence disk.
We found that, in simple yet intuitive examples, using negative momentum makes convergence to the Nash Equilibrium easier. 
Our experiments support the use of negative momentum for saturating losses on mixtures of Gaussians, as well as on other tasks using CIFAR-10 and fashion MNIST. Altogether, fully stabilizing learning in GANs requires a deep understanding of the underlying highly non-linear dynamics. We believe our work is a step towards a better understanding of these dynamics. We encourage deep learning researchers and practitioners to include negative values of momentum in their hyper-parameter search.

We believe that our results explain a decreasing trend in momentum values used for training GANs in the past few years reported in Fig.~\ref{fig:effect}. Some of the most successful papers use zero momentum \citep{arjovsky2017wasserstein,gulrajani2017improved} for architectures that would otherwise call for high momentum values in a non-adversarial setting.

\section*{Acknowledgments}
This research was partially supported by the Canada CIFAR AI Chair Program, the FRQNT nouveaux chercheurs program, 2019-NC-257943, the Canada Excellence Research Chair in ``Data Science for Real-time Decision-making'', by the NSERC Discovery Grant RGPIN-2017-06936, a Google Focused Research Award and an IVADO grant. 
Authors would like to thank NVIDIA corporation for providing the NVIDIA DGX-1 used for this research. Authors are also grateful to Guojun Zhang, Frédéric Bastien, Florian Bordes, Adam Beberg, Cam Moore and Nithya Natesan for their support.

\bibliographystyle{abbrvnat}
\bibliography{negative_momentum}
\clearpage

\appendix
\onecolumn
\section{ADDITIONNAL FIGURES}
\subsection{Maximum magnitude of the eigenvalues gradient descent with negative momentum on a bilinear objective}
\label{sub:plotmag}
\begin{figure}[H]
    \centering
    \includegraphics[width=.48\textwidth]{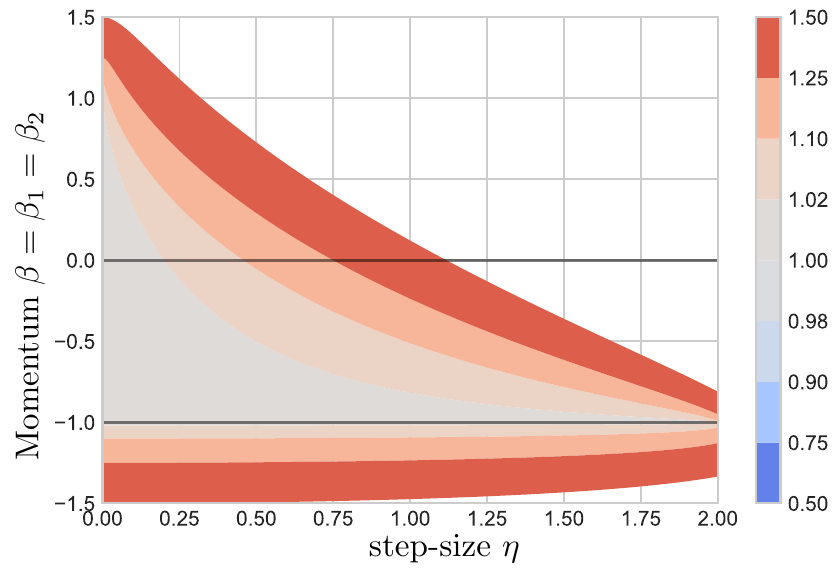}
    \includegraphics[width=.48\textwidth]{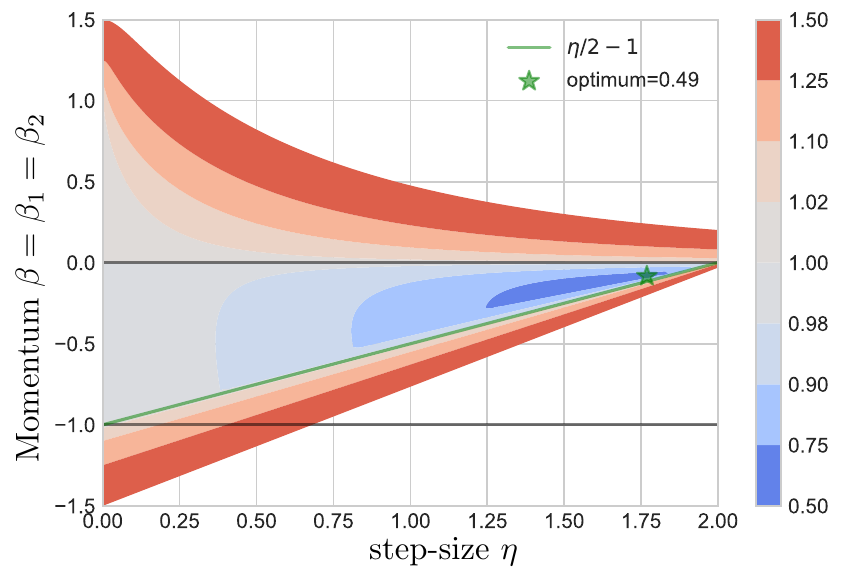}
    \caption{Contour plot of the maximum magnitude of the eigenvalues of
    the polynomial $(x-1)^2(x-\beta)^2 + \eta^2 x^2$ (\textbf{left}, simultaneous)
    and $(x-1)^2(x-\beta)^2 + \eta^2 x^3$ (\textbf{right}, alternated)
    for different values of the step-size $\eta$ and the momentum $\beta$.
    Note that compared to~\eqref{eq:poly_simultaneous} and~\eqref{eq:poly_alternated} 
    we used $\beta_1=\beta_2=\beta$ and we defined $\eta \defas \sqrt{\eta_1 \eta_2 \lambda}$ without loss of generality. 
    On the left,  magnitudes are always larger than $1$, and equal to $1$ for $\beta=-1$.
    On the right, magnitudes are smaller than $1$ for $ \frac{\eta}{2} -1 \leq \beta \leq 0$ and greater than $1$ elsewhere.
    }
    \label{fig:magnitudes}
\end{figure}

In Figure~\ref{fig:magnitudes} we numerically (using the formula provided in Proposition~\ref{prop:eigs_bilinear_sim} and~\ref{prop:eigs_bilinear_alt}) computed the maximum magnitude of the eigenvalues gradient descent with negative momentum on a bilinear objective as a function of the step size $\eta$ and the momentum $\beta$. We can notice that on one hand, for simultaneous gradient method, no value of $\eta$ and $\beta$ provide a maximum magnitude smaller than 1, causing a divergence of the algorithm. On the other hand, for alternating gradient method there exists a sweet spot where the maximum magnitude of the eigenvalues of the operator is smaller than 1 insuring that this method does converge linearly (since the Jacobian of a bilinear minmax proble is constant).

\subsection{Mixture of Gaussian\hspace{-2mm}}
\begin{figure}[H]
  \centering
      \includegraphics[width=0.45\textwidth]{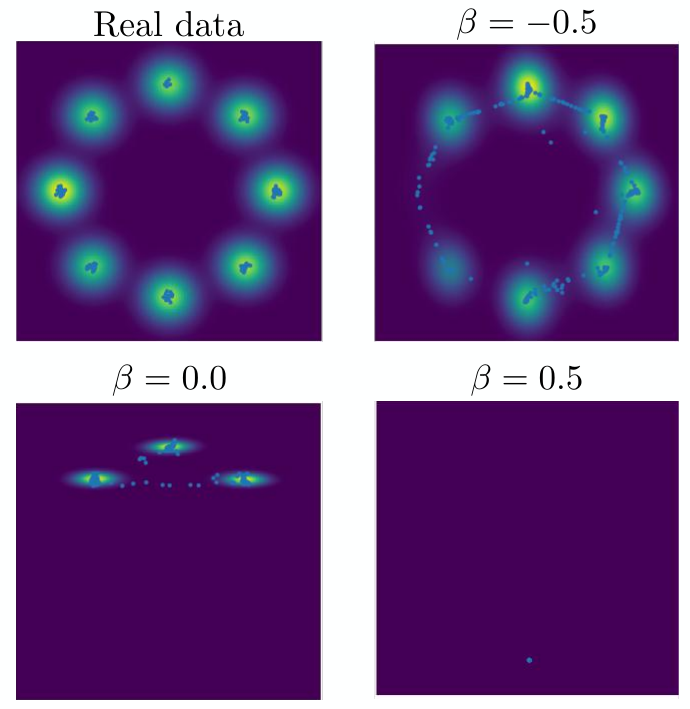}
      \caption{The effect of negative momentum for a mixture of 8 Gaussian distributions in a GAN setup. Real data and the results of using SGD with zero momentum on the Generator and using negative / zero / positive momentum ($\beta$) on the Discriminator are depicted.
}
  \label{fig:8_gauss}
\end{figure}

[Fig.~\ref{fig:8_gauss}]\hspace{2mm}
\label{sub:mog}
In this set of experiments we evaluate the effect of using negative momentum for a GAN with \emph{saturating loss} and alternating steps. The data in this experiment comes from eight Gaussian distributions which are distributed uniformly around the unit circle. The goal is to force the generator to generate 2-D samples that are coming from \emph{all} of the 8 distributions. Although this looks like a simple task, many GANs fail to generate diverse samples in this setup. This experiment shows whether the algorithm prevents mode collapse or not.

We use a fully connected network with 4 hidden \emph{ReLU} layers where each layer has 256 hidden units. The latent code of the generator is an 8-dimensional multivariate Gaussian. The model is trained for 100,000 iterations with a learning rate of $0.01$ for stochastic gradient descent along with values of zero, $-0.5$ and $0.5$ momentum. We observe that negative momentum considerably improves the results compared to positive or zero momentum.

\section{DISCUSSION ON MOMENTUM AND CONDITIONING}\label{sec:kappa_alpha}
In this section, we analyze the effect of the conditioning of the problem on the optimal value of momentum. Consider the following formulation as an extension of the bilinear min-max game discussed in \S\ref{sec:bilinear_game}, Eq. \ref{eq:convex_adversarial_cooperative_games} ($p=d=n$),

\begin{equation} \label{eq:adv_vs_coop_kappa}
    \min_{\vtheta \in \mathbb{R}^n} \max_{\vphi \in \mathbb{R}^n} \; \alpha  \|\mD^{1/2} \vtheta\|_2^2 + (1- \alpha) \vtheta^\top \mA \vphi - \alpha \|\mD^{1/2} \vphi\|_2^2 \, , \quad \alpha \in [0,1] , \; \mA \in \mathbb{R}^{n \times n} \,,
\end{equation}
where $\mD$ is a square diagonal positive-definite matrix,
\begin{equation}
    \mD = \begin{bmatrix}
    d_{1,1} & 0 & 0 & . & . & . & 0 \\
    0 & d_{2,2} & 0 & . & . & . & 0 \\
    0 & 0 & d_{3,3} & . & . & . & 0 \\
    0 & 0 & 0 & . & . & . & 0 \\
    0 & 0 & 0 & . & . & . & d_{n,n} \\
    \end{bmatrix} \text{and}\  \forall j \in \{1, n-1\}, \ d_{{j+1},{j+1}} \geq  d_{j,j} > 0,
\end{equation}
and its condition number is $\kappa (\mD) =d_{n, n} / d_{1,1}$. Thus, we can re-write the vector field and the Jacobian as a function of $\alpha$ and $\mD$,
\begin{equation}
\bm{v}(\vphi, \vtheta, \bm{\alpha}, \mD) =
\begin{bmatrix}
    - (1 - \alpha) \vtheta + 2 \alpha \mD \vphi\\
    2 \alpha \mD \vtheta + (1- \alpha) \vphi
\end{bmatrix},
 \quad\;\;
\nabla \bm{v}(\vphi, \vtheta, \bm{\alpha}, \mD) =
    \begin{bmatrix}
    2 \alpha \mD & (\alpha -1)  \mI_{n} \\
    (1 - \alpha) \mI_{n} & 2 \alpha \mD
\end{bmatrix}.
\end{equation}

The corresponding eigenvalues $\lambda$ of the Jacobian are,
\begin{equation}
    \lambda = 2 \alpha\  d_{j,j} \pm (1-\alpha)i.
\end{equation}
For simplicity, in the following we will note $\nabla F_{\eta, \beta}$ for $\nabla F_{\eta, \beta}(\vphi, \vtheta, \bm{\alpha}, \mD)$.

Using Thm. \eqref{thm:eigen_F_eta_beta}, the eigenvalues of $\nabla F_{\eta, \beta}$ are,
\begin{equation}\label{app:eq:exact}
   \mu_{+}(\beta,\eta,\lambda) = (1 -\eta \lambda + \beta)\frac{1 + \Delta^{\frac{1}{2}}}{2}
   \quad \text{and} \quad
   \mu_{-}(\beta,\eta,\lambda) = (1 -\eta \lambda + \beta)\frac{1 - \Delta^{\frac{1}{2}}}{2} \, .
\end{equation}
where $\Delta := 1 - \frac{4 \beta}{(1 - \eta \lambda + \beta)^2}$ and $\Delta^{\frac{1}{2}}$ is the \emph{complex} square root of $\Delta$ with positive real part. 

Hence the spectral radius of $\nabla F_{\eta, \beta}$ can be explicitly formulated as a function of $\beta$
 and $\eta$,
 \begin{equation}
    \rho(\nabla F_{\eta, \beta}) = \max_{\lambda \in \Sp(\nabla F_{\eta, \beta})} \max \left\{ |\mu_{+}(\beta,\eta,\lambda)|,|\mu_{-}(\beta,\eta,\lambda)| \right\}
\end{equation}

\begin{figure}
  \centering
      \includegraphics[width=0.46\textwidth]{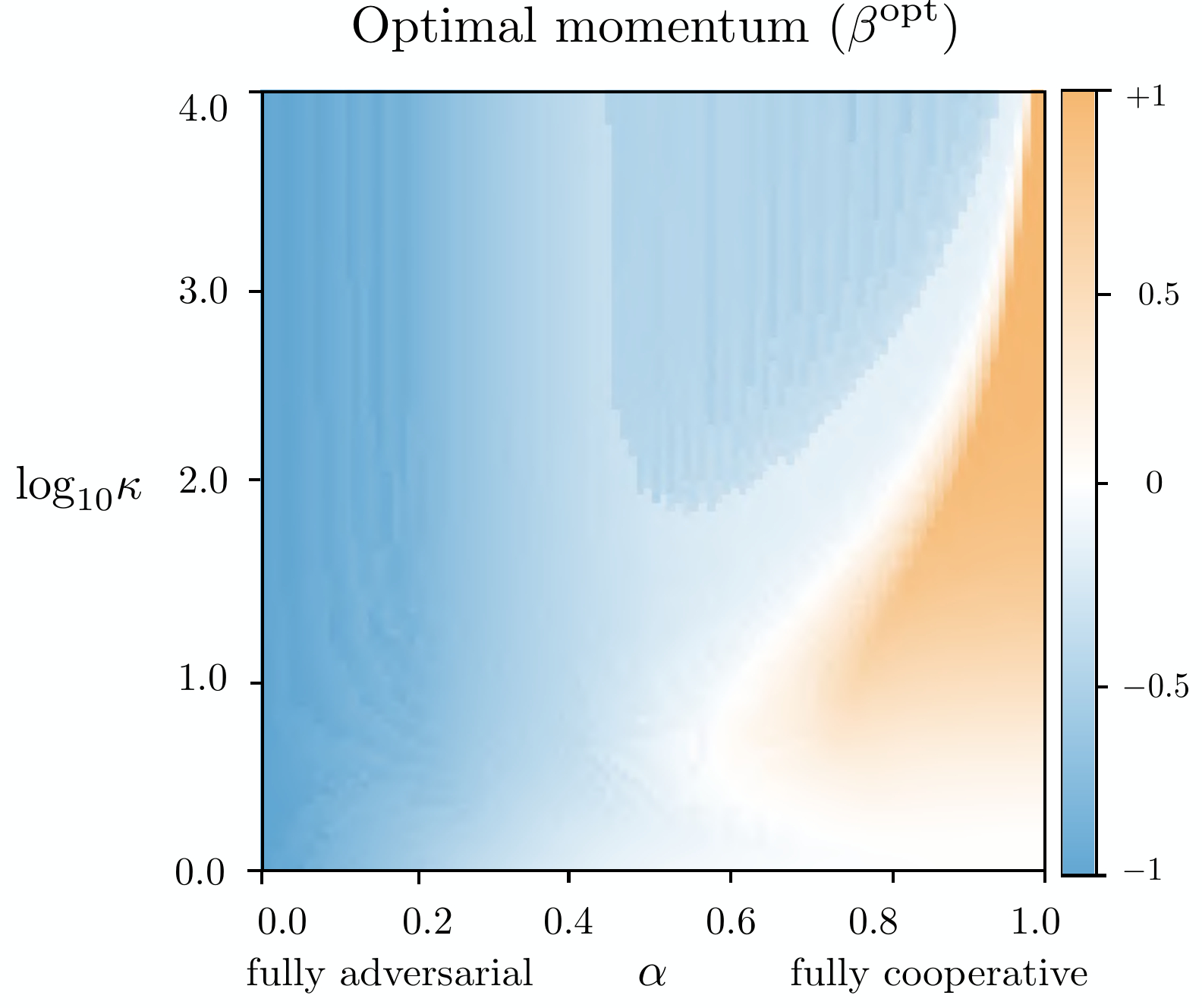}
      \caption{Plot of the optimal value of momentum by for different $\alpha$'s and condition numbers ($log_{10} \kappa$). Blue/white/orange regions correspond to negative/zero/positive values of the optimal momentum, respectively.}
  \label{fig:kappa_alpha}
  \vspace{-2ex}
\end{figure}

In Figure \ref{fig:kappa_alpha}, we numerically computed the optimal $\beta$ that minimizes $\rho_{max}(\nabla F_{\eta, \beta})$ as a function of the step-size $\eta$, for $n = 2$, $d_{1,1}=1 / \kappa$ and $d_{2,2}=1$. To balance the game between the adversarial part and the cooperative part, we normalize the matrix $\mD$ such that the sum of its diagonal elements is $n$. It can be seen that there is a competition between the type of the game (adversarial and cooperative) versus the conditioning of the matrix $\mD$. In a more cooperative regime, increasing $\kappa$ results in more positive values of momentum which is consistent with the intuition that  cooperative games are almost minimization problems where the optimum value for the momentum is known~\citep{polyak1964some} to be $\beta = \big(\frac{\sqrt{\kappa}-1}{\sqrt{\kappa}+1}\big)^2$. Interestingly, even if the condition number of $\mD$ is large, when the game is adversarial enough, the optimum value for the momentum is negative. This experimental setting seems to suggest the existence of a multidimensional condition number taking into account the difficulties introduced by the ill conditioning of $\mD$ as well as the adversarial component of the game.

\section{LEMMAS AND DEFINITIONS}
\label{sec:lemmas_and_definitions}
Recall that the spectral radius $\rho(A)$ of a matrix $A$ is the maximum magnitude of its eigenvalues.
\begin{equation}
    \rho(A) := \max \{ |\lambda| \; : \; \lambda \in \Sp(A)\} \,.
\end{equation}
For a symmetric matrix, this is equal to the spectral norm, which is the operator norm induced by the vector 2-norm. However, we are dealing with general matrices, so these two values may be different. The spectral radius is always smaller than the spectral norm, but it's not a norm itself, as illustrated by the example below:
\begin{equation*}
    \text{If }\mA = 
    \begin{bmatrix}
        0 & 1 \\ 0 & 0
    \end{bmatrix}
    \text{  then  } \bigg( \Sp(\mA)= \{0\} \implies \rho(\mA) = 0 \bigg)
    \text{  but  } \bigg( \mA^\top \mA = 
    \begin{bmatrix}
        1 & 0 \\ 0 & 0
    \end{bmatrix}
    \implies \| \mA \|_2 = 1 \bigg)
\end{equation*}
where we used the fact that the spectral norm is also the square root of the largest singular value.

In this section we will introduce three lemmas that we will use in the proofs of \S\ref{sec:proof_of_theorems}.

The first lemma is about the determinant of a block matrix.
\begin{lemma} \label{lemma:block_det}
Let $A,B,C,D$ four matrices such that $C$ and $D$ commute. Then 
\begin{equation}
  \begin{vmatrix}
  A & B \\ C & D
  \end{vmatrix} 
  = \begin{vmatrix}
  AD - BC
  \end{vmatrix}
\end{equation}
  where $\begin{vmatrix} A \end{vmatrix}$ is the determinant of $A$.
\end{lemma}
\proof 
See~\citep[Section 0.3]{zhang2006schur}.
\endproof

The second lemma is about the iterates of the simultaneous and the alternating methods introduced in \S\ref{sec:bilinear_game}  for the bilinear game.
It shows that we can pick a subspace where the iterates will remain.

\begin{lemma}\label{lemma:reduc_dim}
    Let $(\vtheta_t,\vphi_t)$ the updates computed by the simultaneous (resp. alternating) gradient method with momentum~\eqref{eq:updates_sim} (resp.~\eqref{eq:updates_alt}). There exists are couple $(\vtheta^*,\vphi^*)$ solution of~\eqref{eq:adversarial_games} only depending on $(\vtheta_0,\vphi_0)$ such that,
    \begin{equation}\label{eq:belong_span_1}
    \vtheta_t - \vtheta^* \in span(\mA \mA^\top) \quad \text{and} \quad \vphi_t - \vphi^* \in span(\mA^\top \mA) \, , \quad \forall t\geq 0\,.
 \end{equation}
    \end{lemma}
\proof[\textbf{Proof of Lemma~\ref{lemma:reduc_dim}}]
Let us start with the simultaneous updates~\eqref{eq:updates_sim}.

Let $\mU^\top \mD \mV = \mA$ the SVD of $\mA$ where $\mU$ and $\mV$ are orthogonal matrices and
\begin{equation}
    D = \begin{bmatrix}
        \diag(\sigma_1, \ldots,\sigma_r) & \bm{0}_{r,p-r} \\
        \bm{0}_{d-r,r} & \bm{0}_{d-r,p-r}
    \end{bmatrix}
\end{equation}
where $r$ is the rank of $A$ and $\sigma_1 \geq \cdots \geq \sigma_r>0$ are the (positive) singular values of $A$.
The update rules~\eqref{eq:updates_sim} implies that,
\begin{equation}\label{eq:update_ortho}
    \left\{
    \begin{aligned}
    \vtheta_{t+1} &= \vtheta_t - \eta_1 \mA \vphi_t + \beta_1(\vtheta_t-\vtheta_{t-1}) \\
        \vphi_{t+1} &= \vphi_t + \eta_2 \mA^\top \vtheta_t + \beta_2 (\vphi_t - \vphi_{t-1})
    \end{aligned}
    \right.
    \Rightarrow
    \left\{
    \begin{aligned}
    \mU\vtheta_{t+1} &= \mU\vtheta_t - \eta_1 \mD \mV\vphi_t + \beta_1\mU(\vtheta_t-\vtheta_{t-1}) \\
     \mV   \vphi_{t+1} &= \mV\vphi_t + \eta_2 \mD^\top \mU\vtheta_t + \beta_2 \mV (\vphi_t - \vphi_{t-1})
    \end{aligned}
    \right.
\end{equation}
Consequently, for any $\vtheta_0 \in \R^d$ and $\vphi_0 \in \R^p$ we have that,
\begin{equation}\label{eq:theta_phi_0_ker}
    \mA^\top \left( \mU^\top \begin{bmatrix}
        0 \\ \vdots \\ 0 \\ [\mU \vtheta_0]_{r+1} \\ \vdots \\ [\mU \vtheta_0]_{d}
    \end{bmatrix} \right) =  \bm{0}
    \quad \text{and} \quad 
     \mA \left( \mV^\top \begin{bmatrix}
        0 \\ \vdots \\ 0 \\ [\mV \vphi_0]_{r+1} \\ \vdots \\ [\mV \vphi_0]_{d}
    \end{bmatrix} \right) =  \bm{0}
\end{equation}
Since the solutions $(\vtheta^*,\vphi^*)$ of \eqref{eq:adversarial_games}
verify the following first order conditions: 
\begin{equation}
      \mA^\top \vtheta^* =  \bm{0}
    \quad \text{and} \quad 
     \mA \vphi^* =  \bm{0}
\end{equation}
 One can set $(\vtheta^*,\vphi^*)$ as in \eqref{eq:theta_phi_0_ker} to be a couple of solution of \eqref{eq:adversarial_games} such that $\mU (\vtheta_0 - \vtheta^*) \in span(D)$ and $\mV (\vphi_0-\vphi^*)\in span(D)$. By an immediate recurrence, using \eqref{eq:update_ortho} we have that for any initialization $(\vtheta_0,\vphi_0)$ there exists a couple $(\vtheta^*,\vphi^*)$ such that that for any $t \geq 0$, 
 \begin{equation}\label{eq:belong_span_2}
     \mU (\vtheta_t - \vtheta^*) \in span(\mD) \quad \text{and} \quad \mV(\vphi_t - \vphi^*) \in span(\mD^\top) 
 \end{equation}
 Consequently,
 \begin{equation}
     \vtheta_t - \vtheta^* \in span(\mA)= span(\mA\mA^\top) \quad \text{and} \quad \vphi_t - \vphi^* \in span(\mA^\top) = span(\mA^\top \mA) \, , \quad t \geq 0 \,
 \end{equation}
 The proof for the alternated updates~\eqref{eq:updates_alt} are the same since we only use the fact that the iterates stay on the span of interest.
\endproof

\begin{lemma}\label{lemma:rate_eigs}
    Let $\mM \in \R^{m\times m}$ and $(\vu_t)$ a sequence such that, $\vu_{t+1} = \mM \vu_t$, then we have three cases of interest for the spectral radius $\rho(M)$: 
    \begin{itemize}
        \item If $\rho(M)<1$, and $M$ is diagonalizable, then $\|\vu_t\|_2 \in O( (\rho(\mM))^t\|\vu_0\|_2)$.
        \item If $\rho(M)>1$, then there exist $\vu_0$ such that $\|\vu_t\|_2 \in \Omega (\rho(\mM))^t\|\vu_0\|_2$.
        \item If $|\lambda| = 1 \, , \; \forall \lambda \in Sp(M)$, and $M$ is diagonalizable then $\|\vu_t\|_2 \in \Theta (\|\vu_0\|_2)$.
    \end{itemize}
\end{lemma}
\proof
For that section we note $\|\cdot\|_2$ the $\ell_2$ norm of $\mathbb{C}^m$:
\begin{itemize}
    \item If $\rho(M)<1$:
    
    We have for $t \geq 0$ and any $\vu_0 \in \R^m$ ,
\begin{equation}
    \|\vu_t\|_2 = \|\mM^t\vu_0\|_2 \leq \|\mM^t\| \|\vu_0\|_2 
\end{equation}
Then we can diagonalize $\mM = \mP \mD \mP^{-1}$ where $\mP$ is invertible and $\mD$ is a diagonal matrix. Hence using $\|\cdot\|_2$ as the norm of $\mathbb{C}^m$ (because $\mP$ can belong to $\mathbb{C}^{m\times m}$) we have that,
\begin{equation}
    \|\vu_t\|_2 \|\mP \mD^t\mP^{-1}\| \|\vu_0\|_2  \leq \|\mP\|\|\mP^{-1}\|\|\mD^t\| \|\vu_0\|_2 \leq \|\mP\|\|\mP^{-1}\| \rho(M)^t \|\vu_0\|_2 = O( (\rho(\mM))^t\|\vu_0\|_2) \,. 
\end{equation}
    \item If $\rho(M)>1$:
    We have for $t \geq 0$ and any $\vu_0 \in \R^m$ ,
\begin{equation}
    \|\vu_t\|_2 = \|\mM^t\vu_0\|_2
\end{equation}
But we know that there exist a $\vu_0 \in \R^m$ that only depends on $M$ such that $\|\mM^t\vu_0\|_2 = \|\mM^t\|\|\vu_0\|_2$ (explicitly $\vu_0$ is the eigenvector associated with the largest eigenvalue of $\mM^\top \mM$).
But, using \citep[Proposition A.15]{bertsekas1999nonlinear} we know that $\rho(\mM)\leq \|\mM\|_2$. Then we have that,
\begin{equation}
    \|\vu_t\|_2  \geq \rho(\mM)^t \|\vu_0\|_2 
\end{equation}
\item If $|\lambda| =1 \, , \; \forall \lambda \in Sp(M)$, we can diagonalize $\mM$ such that $\mM = \mP \mD \mP^{-1}$ where $\mP$ is invertible and $\mD$ is a diagonal matrix with complex values of magnitude 1.  

 We have for $t \geq 0$ and any $\vu_0 \in \R^m$,
\begin{equation}
    \|\vu_t\|_2 = \|\mM^t\vu_0\|_2 = \|\mP \mD^t \mP^{-1}\vu_0\|_2 \leq \|\mP\| \|\mD^t\|\|\mP^{-1}\| \|\vu_0\|_2 = \|\mP\|\|\mP^{-1}\| \|\vu_0\|_2
\end{equation}
Similarly, 
\begin{equation}
    \|\vu_0\|_2 = \|\mM^{-t}\vu_t\|_2 = \|\mP \mD^{-t} \mP^{-1}\vu_t\|_2 \leq \|\mP\| \|\mD^t\|\|\mP^{-1}\| \|\vu_t\|_2 = \|\mP\|\|\mP^{-1}\| \|\vu_t\|_2
\end{equation}
\end{itemize}

\endproof

\section{PROOFS OF THE THEOREMS AND PROPOSITIONS}
\label{sec:proof_of_theorems}

\subsection{Proof of Thm.~\ref{thm:contractive_operator}}
\label{sub:contractive_operator}

Let us recall the Theorem proposed by~\citet[Proposition 4.4.1]{bertsekas1999nonlinear}. We also provide a convergence rate that was not previously stated in~\citep{bertsekas1999nonlinear}.
\begin{reptheorem}{thm:contractive_operator}
\thmContractiveOperator
\end{reptheorem}

\proof
For brevity let us write $x_t := (\phi_t,\theta_t)$ for $t \geq 0$ and $x^* := (\phi^*,\theta^*) $. Let $\epsilon >0$.

By Proposition A.15~\citep{bertsekas1999nonlinear} there exists a norm $\|\cdot\|$ such that its induced matrix norm has the following property: 
\begin{equation} \label{eq:induced_norm_close_t_spectral_radius}
    \|\nabla F_\eta (x^*)\| \leq \rho(\nabla F_\eta (x^*)) + \frac{\epsilon}{2} \,. 
\end{equation}
Then by definition of the sequence $(x_t)$ and since $x^*$ is a fixed point of $F_\eta$, we have that,
\begin{equation}
    \|x_{t+1} - x^*\| = \|F_\eta(x_t) - F_\eta(x^*)\|
\end{equation}
Since $F_\eta$ is assumed to be continuously differentiable by the mean value theorem we have that
\begin{equation}
    F_\eta(x_t) = F_\eta(x^*) + \nabla F_\eta (\tilde{x}_t) (x_t-x^*) \,,
\end{equation}
for some $\tilde{x}_t \in [x_t,x^*]$. Then,
\begin{equation}
    \|x_{t+1} - x^*\| \leq \|\nabla F_\eta (\tilde{x}_t)\| \|x_{t} - x^*\|
\end{equation}
where $\|\nabla F_\eta (\tilde{x}_t)\|$ is the induced matrix norm of $\|\cdot\|$. 

Since the induced norm of a square matrix is continuous on its elements and since we assumed that $\nabla F_\eta$ was continuous, there exists $\delta>0$ such that,
\begin{equation}\label{eq:nabla_continuous}
    \|\nabla F_\eta (x) - \nabla F_\eta (x^*)\| \leq \frac{\epsilon}{2} \,, \quad \forall x \;:\; \|x - x^*\|\leq \delta \, .
\end{equation}
Finally, we get that if $\|x_t - x^*\|\leq \delta$, then,
\begin{align}
    \|x_{t+1} - x^*\| 
    & \leq \|\nabla F_\eta (\tilde{x}_t)\| \|x_{t} - x^*\| \\
    & \leq \left(  \|\nabla F_\eta (x^*)\| +  \|\nabla F_\eta (\tilde{x}_t) - \nabla F_\eta (x^*)\| \right) \|x_{t} - x^*\| \\
    & \leq \left( \rho(\nabla F_\eta (x^*)) + \frac{\epsilon}{2}+ \frac{\epsilon}{2} \right)\|x_{t} - x^*\|
\end{align}
where in the last line we used~\eqref{eq:induced_norm_close_t_spectral_radius} and~\eqref{eq:nabla_continuous}. Consequently, if $ \rho(\nabla F_\eta (x^*)<1$ and if $\|x_0 - x^*\| \leq \delta$, we have that,
\begin{equation}
    \|x_{t} - x^*\| \leq \left( \rho(\nabla F_\eta (x^*)) + \epsilon \right)^t \|x_0 - x^*\| \leq  \delta  \,, \quad  \forall \epsilon >0\,.
\end{equation}

\endproof

\subsection{Proof of Thm.~\ref{thm:best_step_size}}
\label{sub:thm:best_step_size}

We are interested in the optimal step-size for the Simultaneous gradient with no momentum.
Define the step-size associated to one eigenvalue $\lambda \in \mathbb{C}$ by $\hat \eta(\lambda) \defas \frac{\Re(\lambda)}{|\lambda|^2}$.

\begin{reptheorem}{thm:best_step_size}
\thmBestStepSize
\end{reptheorem}

\proof

The eigenvalues of $\nabla F_{\eta}$ are $1 - \eta \lambda$, for $\lambda \in \Sp(\nabla \bm{v}(\vphi,\vtheta))$. Our goal is to solve
\begin{equation}
    \rho_\text{max} \defas \min_{\eta \geq 0} \max_{1 \leq i \leq m} |1 - \eta \lambda_i|^2
\end{equation}
where $\{\lambda_1,\ldots,\lambda_m\}$ is the spectrum of $\nabla \bm{v}(\vphi^*,\vtheta^*)$.
we can develop the magnitude to get,
\begin{equation}
    f_i(\eta) := |1 - \eta \lambda_i|^2 = 1 - 2 \eta \Re(\lambda_i) + \eta^2 |\lambda_i|^2
\end{equation}
The function $\eta \mapsto \max_{1 \leq i \leq n} f_i(\eta)$ is a convex function quadratic by part. 
This function goes to $+\infty$ as $\eta$ gets larger, so it reaches its minimum over $[0,\infty)$.
We can notice that each function $f_i$ reaches its minimum for $\eta_i = \frac{\Re(\lambda_i)}{|\lambda_i|^2}= \Re(1/\lambda_i)$. 
Consequently, if we order the eigenvalues such that,
\begin{equation}
   \eta_1 \leq \ldots \leq \eta_m
\end{equation}
we have that 
\begin{equation}
    f'_i(\eta_1) \leq 0 
     \, ,\quad 1\leq i \leq m
     \quad
     \text{and}
     \quad 
    f_1(x) \geq 1 \, , \, \forall x \geq 2 \eta_1
\end{equation} 
As a result, 
\begin{equation}
    \eta_1 \leq \eta_{best} \leq 2\eta_1
\end{equation}
Moreover, it is easy to notice that, 
\begin{equation}
  |1 - \eta_1 \lambda_1|^2 = \min_{\eta \geq 0} |1 - \eta \lambda_1|^2 \leq \min_{\eta \geq 0} \max_{1 \leq k \leq m} |1 - \eta \lambda_i|^2  
\end{equation}
Then developing $|1 - \eta_1   \lambda_1|^2$, we get that,
\begin{equation}
    |1 - \eta_1   \lambda_1|^2 = |1 - \tfrac{\Re(\lambda_1)}{|\lambda_1|^2}  \lambda_1|^2 = 1 - \tfrac{\Re(\lambda_1)^2}{|\lambda_1|^2} = \sin(\psi_1)^2 \label{eq:min_mag}
\end{equation}
where $\lambda_1 = r_1 e^{i \psi_1}$. Moreover, we also have that
\begin{equation}
  \rho_{\max} = \min_{\eta \geq 0} \max_{1 \leq i \leq n} |1 - \eta \lambda_i|^2  
  \leq \max_{1 \leq k \leq m} |1 - \eta_1 \lambda_i|^2  = 1 - \eta_1  \min_{1 \leq k \leq m} 2 \Re(\lambda_k) - \eta_1 |\lambda_k|^2
  = 1 - \Re(1/\lambda_1) \delta
\end{equation}
This upper bound is then achieved for $\eta = \Re(1/\lambda_1)$. Moreover is $\Sp(\nabla \vv (\vphi^*,\vtheta^*) \subset [\mu,L]$ we have that, $\lambda_1 = L$ and that
\begin{equation}
    \delta \geq \min_{\lambda \in [\mu,L]} 2 \lambda -  \lambda^2/L = 
    2 \mu - \mu^2/L
\end{equation}
Consequently we recover the standard upper bound $\rho_{\max}^2 \leq 1 - 2 \frac{\mu}{L} + \frac{\mu^2}{L}= (1 - \mu/L)^2$ provided in the convex case.

\endproof

\subsection{Proof of Thm.~\ref{thm:eigen_F_eta_beta}}
\label{sub:thm:eigen_F}
We are now interested in the eigenvalues of the Simultaneous Gradient Method with Momentum.

\begin{reptheorem}{thm:eigen_F_eta_beta}
\thmEigenMomentum
\end{reptheorem}

\proof
The Jacobian of $F_{\eta,\beta}$ is 
\begin{equation}
    M := \begin{bmatrix}
    \bm{I}_n - \eta \nabla\bm{v}(\vomega^*) + \beta \bm{I}_n & -\beta \bm{I}_n \\ \bm{I}_n & \bm{0}_n
  \end{bmatrix}
\end{equation}
Its characteristic polynomial can be written:
\begin{equation}
  \chi_M (X) = \det(X \bm{I}_{2n} - M) = \begin{vmatrix}
    (X - 1 - \beta) \bm{I}_n + \eta T& \beta \bm{I}_n \\ -\bm{I}_n & X \bm{I}_n
  \end{vmatrix}
\end{equation}
where $\nabla\bm{v}(\vomega^*) =  PTP^{-1}$ and $T$ is an upper-triangular matrix. 
Finally by Lemma~\ref{lemma:block_det} we have that,
\begin{equation}
  \chi_M (X) 
  = \begin{vmatrix}  X((X - 1 - \beta) \bm{I}_n + \eta T) +\beta \bm{I}_n  \end{vmatrix} 
  = \prod_{i=1}^n \left( X((X - 1 - \beta) + \eta \lambda_i) +\beta \right) 
\end{equation}
where 
$$T = \begin{bmatrix} 
          \lambda_1 & * & \ldots  & * \\ 
          0 & \ddots & \ddots  & \vdots \\
          \vdots & \ddots & \ddots& * \\
          0 & \ldots & 0 & \lambda_{n}
   \end{bmatrix} \; .$$
Let $\lambda$ one of the $\lambda_i$ we have,
\begin{equation}
   X((X - 1 - \beta) + \eta \lambda) +\beta = X^2 - (1 -  \eta \lambda + \beta)X + \beta
\end{equation}
The roots of this polynomial are 
\begin{equation}
  \mu_+(\lambda) = \frac{ 1 - \eta \lambda + \beta + \sqrt{\Delta}}{2} 
  \quad \text{and} \quad
  \mu_-(\lambda) = \frac{ 1 - \eta \lambda + \beta - \sqrt{\Delta}}{2}
\end{equation}
where $\Delta := (1 - \eta \lambda + \beta)^2 - 4 \beta$
and $\lambda \in \Sp(\nabla\bm{v}(\vomega^*))$. This can be rewritten as, 
\begin{equation}
   \mu_{\pm}(\beta,\eta,\lambda):= (1 -\eta \lambda + \beta)\frac{1 \pm \Delta^{\frac{1}{2}}}{2}
\end{equation}
where $ \Delta := 1 - \frac{4 \beta}{(1 - \eta \lambda + \beta)^2}  \, , \; \lambda \in \Sp(\nabla\bm{v}(\vphi^*,\vtheta^*))$ and $\Delta^{\frac{1}{2}}$ is the \emph{complex} square root of $\Delta$ with real positive part (if $\Delta$ is a real negative number, we set $\Delta^{\frac{1}{2}}:= i \sqrt{-\Delta}$). Moreover we have the following Taylor approximation,
\begin{equation}\label{app:eq:asymptotic}
   \mu_{+}(\beta,\eta,\lambda) = 1 - \eta \lambda - \beta \frac{\eta \lambda}{1 - \eta \lambda} + O(\beta^2) 
   \quad \text{and} \quad
   \mu_{-}(\beta,\eta,\lambda) = \frac{\beta}{1-\eta\lambda} + O(\beta^2).
\end{equation}
\endproof

\subsection{Proof of Thm.~\ref{thm:derivative_positive}}
\label{sub:thm:derivative_positive}
We are interested in the impact of small Momentum values on the convergence rate of Simultaneous Gradient Method.

\begin{reptheorem}{thm:derivative_positive}
    \thmDerivativePositive
\end{reptheorem}

\proof
Recall the definitions of $\mu_+$ and $\mu_-$ from Thm.~\ref{thm:eigen_F_eta_beta}, and the definition of the radius:
\begin{equation}
  \rho_{\lambda,\eta}(\beta) 
  := \max \left\{|\mu_+|^2, |\mu_-|^2 \right\}
\end{equation}

When $\beta$ is close to $0$, $\mu_-$ is close also to $0$ whereas $\mu_+$ is close to $1 - \eta \lambda$.
In general $1 - \eta \lambda \neq 0$, 
so around $0$, $\rho_{\lambda,\eta}(\beta) = |\mu_+(\beta)|^2 = \mu_+(\beta) \bar \mu_+(\beta)$.
The special case where $1 - \eta \lambda = 0$ is excluded from this analysis because it means that the eigenvalue $\lambda$ is not one constraining the learning rate as seen in Thm.~\ref{thm:best_step_size}.
Computing the derivative of $\rho$ give us
\begin{align}
    \rho'_{\lambda,\eta}(0) 
    & = (\mu_+ \bar \mu_+)'(0) = \mu_+(0) \bar \mu_+'(0) + \bar \mu_+(0) \mu_+'(0) \\
    & = 2 \Re(\mu_+(0) \bar \mu_+'(0)) \\
    & = 2 \Re \left( (1 -\eta \lambda) \frac{- \eta \bar \lambda }{1 -\eta \bar \lambda} \right) \\
    & = - 2 \eta \Re\left( \frac{ \bar{\lambda}(1 -\eta \lambda)^2}{|1 -\eta \lambda|^2} \right) \\
    & = \frac{-2 \eta}{|1-\eta \lambda|^2} \left[ \Re(\lambda) - 2 \eta|\lambda|^2 + \eta^2 |\lambda|^2 \Re(\lambda ) \right]
\end{align}
which leads to,
\begin{equation}
     \rho'_{\lambda,\eta}(0) = 2 \frac{2\eta^2|\lambda|^2 - \eta \Re(\lambda)(1+ \eta^2|\lambda|^2)}{|1-\eta \lambda|^2} 
\end{equation}
The sign of $\rho'_{\lambda,\eta}(0)$ is determined by the sign of
\begin{equation}
    2\eta|\lambda|^2 - \Re(\lambda)(1+ \eta^2|\lambda|^2) 
    = - \Re(\lambda)|\lambda|^2 \eta^2 + 2|\lambda|^2 \eta - \Re(\lambda)
\end{equation}
This quadratic function is strictly positive on the open interval $\left(\frac{|\lambda|-|\Im(\lambda)|}{|\lambda| \Re(\lambda)}, \frac{|\lambda|+| \Im(\lambda)|}{|\lambda|\Re(\lambda)}\right)$.

Moreover since $\Re(1/\lambda) = \frac{\Re(\lambda)}{|\lambda|^2}$, we have that $|1 - \lambda \Re(1/\lambda)|^2 = 1 - \Re(\lambda)\Re(1/\lambda)$ (see Eq.~\ref{eq:min_mag}) and then,
\begin{equation}
\rho'_{\lambda,\Re(1/\lambda)}(0) = 2 \Re(\lambda)\Re(1/\lambda).    
\end{equation}
Finally writting $\lambda = r e^{i \psi}$ we get that, 
\begin{equation}
     \frac{|\lambda|-|\Im(\lambda)|}{|\lambda| \Re(\lambda)}
     = \frac{1 - |\sin(\psi)|}{r \cos(\psi)} = \Re(1/\lambda) \frac{1 - |\sin(\psi)|}{1 - |\sin(\psi)|^2}
     \quad \text{and} \quad 
     \frac{|\lambda|+| \Im(\lambda)|}{|\lambda|\Re(\lambda)} 
     = \frac{1 + |\sin(\psi)|}{r \cos(\psi)}
     =  \Re(1/\lambda) \frac{1 + |\sin(\psi)|}{1 - |\sin(\psi)|^2}
     \notag
\end{equation}
Consequently,
\begin{equation}
    I(\lambda) = \left(  \frac{\Re(1/\lambda)}{1 + |\sin(\psi)|},  \frac{\Re(1/\lambda)}{1 - |\sin(\psi)|}\right)
\end{equation}
and $|\arg(\lambda)| \geq \frac{\pi}{4}$ implies that $\left(\frac{2}{3}\Re(1/\lambda),  2\Re(1/\lambda)\right)$
\endproof

\subsection{Proof of Thm.~\ref{thm:bilin_sim}}
\label{sub:thm_bilin_sim}

We are now in the special case of a bilinear game. 
We first consider the simultaneous gradient step with momentum
The operator $F_{\eta,\beta}$ is defined as:
\begin{equation}
    F^{\text{sim}}_{\eta,\beta} \begin{bmatrix} \vtheta_t \\\vphi_t \\\vtheta_{t-1} \\ \vphi_{t-1} \end{bmatrix} \defas \begin{bmatrix}
        \vtheta_t - \eta_1 \mA \vphi_t + \beta_1 (\vtheta_t - \vtheta_{t-1}) \\
        \vphi_t + \eta_2 \mA^\top \vtheta_t+ \beta_2 (\vphi_t - \vphi_{t-1}) \\
        \vtheta_t \\
        \vphi_t
    \end{bmatrix} \; .
\end{equation}

\begin{repproposition}{prop:eigs_bilinear_sim}
  \propEigsSim
  Particularly, when $\beta_1 = \beta_2  = 0$ and $\eta_1=\eta_2 = \eta$ we have,
  \begin{equation}
      P_\lambda(x) = x^2 ( x^2- 2x + 1 + \eta^2 \lambda) \; , \; \lambda \in \Sp(\mA^\top \mA)  
  \end{equation}
\end{repproposition}

\proof
$F^{\text{sim}}_{\eta,\beta}$ is a linear operator belonging to $\R^{d \times p}$, for notational compactness let us call $m \defas d + p$. Let us recall that $\mI_m$ and $\bm{0}_{d,p}$ are respectively the identity of $\R^{m \times m}$ and the zero matrix of $\R^{d\times p}$.
\begin{equation}
\nabla F^{\text{sim}}_{\eta,\beta} = 
  \begin{bmatrix}
    \bm{I}_{m} & \bm{0}_{m}\\ 
    \bm{I}_{m} & \bm{0}_{m} 
    \end{bmatrix}   
    +  \begin{bmatrix}
    \begin{matrix}
    \bm{0}_{d} & -\eta_1\mA \\
     \eta_2\mA^\top & \bm{0}_p        
    \end{matrix}
    & \bm{\bigzero}_m \\[4mm]
\bigzero_{m}&  \bigzero_m 
    \end{bmatrix}   
    +
    \begin{bmatrix}
    \begin{matrix}
    \beta_1\bm{I}_{d} &  \bm{0}_{d,p} \\
    \bm{0}_{p,d} & \beta_2 \bm{I}_{p}
    \end{matrix} &
    \begin{matrix}
    -\beta_1\bm{I}_{d} &  \bm{0}_{d,p} \\
    \bm{0}_{p,d} & -\beta_2 \bm{I}_{p}
    \end{matrix} \\[4mm]
    \bigzero_m & \bigzero_m
    \end{bmatrix} 
\end{equation}
Leading to the compressed form
\begin{equation}
\nabla F^{\text{sim}}_{\eta,\beta} =  
    \begin{bmatrix}
    \begin{matrix}
    (1+\beta)\bm{I}_{d} & -\eta\mA \\
     \eta\mA^\top & (1+\beta)\bm{I}_p        
    \end{matrix} &
    \begin{matrix}
    -\beta_1\bm{I}_{d} &  \bm{0}_{d,p} \\
    \bm{0}_{p,d} & -\beta_2 \bm{I}_{p}
    \end{matrix} &\\[4mm]
\bm{I}_{m}&  \bigzero_m 
    \end{bmatrix}
\end{equation}
Then the characteristic polynomial of this matrix is equal to,
\begin{equation}
\chi(X) := \begin{vmatrix}
    \begin{matrix}
    (X-1-\beta_1)\bm{I}_{d} &  \eta \mA \\
     -\eta \mA^\top & (X-1-\beta_2)\bm{I}_{p}
    \end{matrix}
    & 
    \begin{matrix}
    \beta_1\bm{I}_{d} &  \bm{0}_{d,p} \\
    \bm{0}_{p,d} & \beta_2 \bm{I}_{p}
    \end{matrix} &\\[4mm]  
    -\bm{I}_m &  X\bm{I}_m
    \end{vmatrix} 
\end{equation}
Then we can use Lemma~\ref{lemma:block_det} to compute this determinant,
\begin{align}
\chi(X) &= \det\left( X\begin{bmatrix}
    (X-1-\beta_1)\bm{I}_{d} &  \eta_1 \mA   \\
    -\eta_2 \mA^\top & (X-1-\beta_2)\bm{I}_{p} 
    \end{bmatrix} 
    +
    \begin{bmatrix}
    \beta_1\bm{I}_{d} &  \bm{0}_{d,p} \\
    \bm{0}_{p,d} & \beta_2 \bm{I}_{p}
    \end{bmatrix} \right) \\
    & = \begin{vmatrix}
    (X(X-1-\beta_1) +  \beta_1)\bm{I}_{d} &  \eta_1 X \mA   \\
    -\eta_2 X \mA^\top & (X(X-1-\beta_2)+\beta_2)\bm{I}_{p} 
    \end{vmatrix} \\
    & = 
    \begin{vmatrix}
    (X-\beta_1)(X-1)\bm{I}_{d}  + \eta_1 \eta_2 \frac{X^2}{X(X-1-\beta_2) +  \beta_2} \mA^\top \mA&  \eta_1 X \mA \\
    \bm{0}_{d,p} & (X-\beta_2)(X-1)\bm{I}_p 
    \end{vmatrix} 
\end{align}
Where for the last equality we added to the first block column the second one multiplied by $\eta_2\mA^\top\frac{X}{X(X-1-\beta_2) +  \beta_2}$.
It's now time to introduce $r$ the rank of $\mA$.
We can diagonalize $\mA^\top \mA = \mU^\top diag(\lambda_1,\ldots,\lambda_r,0,\ldots,0)\mU$ to get the determinant of a triangular matrix,
\begin{align}
  \chi(X)
& = ((X-\beta_1)(X-1))^{d-r}((X-\beta_2)(X-1))^{p-r} \prod_{k=1}^r \left[(X-\beta_1)(X-1)(X-\beta_2)(X-1) + \eta_1 \eta_2 X^2\lambda_k \right]
\end{align}
where $\lambda_k$ are the positive eigenvalues of $\mA^\top \mA$.
This is the characteristic polynomial we were seeking, taking into account the null singular values of $A$.

In particular, when $\beta_1 = \beta_2 = 0$, we get, 
\begin{align}
  \chi(X) = X^{m}(X-1)^{m-2r} \prod_{k=1}^r ((X-1)^2+ \eta_1 \eta_2\lambda_k )
\end{align}

\endproof
\begin{reptheorem}{thm:bilin_sim}
    \thmDivBilin
\end{reptheorem}

\proof[\textbf{Proof of Thm.~\ref{thm:bilin_sim}}]
We report the maximum magnitudes of the eigenvalues of the polynomial from Prop.~\ref{prop:eigs_bilinear_sim} in Fig.~\ref{fig:magnitudes}.
We observe that they are larger than 1.
We now prove it in several cases.
Let us start with the simpler case $\beta_1=\beta_2=0$. Using Lemma~\ref{lemma:reduc_dim}, there exists $(\vtheta^*,\vphi^*)$ such that for any $t \geq 0$,
 \begin{equation}\label{eq:belong_span}
     \vtheta_t - \vtheta^* \in span(\mA)= span(\mA\mA^\top) \quad \text{and} \quad \vphi_t - \vphi^* \in span(\mA^\top) = span(\mA^\top \mA) \, , \quad t \geq 0 \,
 \end{equation}
 Then, we have,
 \begin{align}
    \|\vtheta_{t+1}-\vtheta^*\|^2
    &= \|\vtheta_t - \vtheta^* - 2\eta \mA (\vphi_t - \vphi^*) \|^2 \\
    &= \|\vtheta_t - \vtheta^*\|^2 - 2\eta (\vtheta_t - \vtheta^*) \mA(\vphi_t - \vphi^*) + \eta^2 \|\mA(\vphi_t - \vphi^*)\|^2 \\
    & \overset{\eqref{eq:belong_span}}{\geq} \|\vtheta_t - \vtheta^*\|^2 - 2\eta (\vtheta_t - \vtheta^*) \mA(\vphi_t - \vphi^*) + \eta^2 \sigma^2_{\min}(\mA)\| \vphi_t - \vphi^*\|^2  \label{ineq:phi_thm1}
\end{align}
where in line 1 we used that $\mA\vphi^* = 0 $ and in line 3 we used that $\vphi_t - \vphi^*$ is orthogonal to the null space of $\mA$, so that we lower bound the product by the smallest non-zero singular value $\sigma_{\min}(\mA)$. The same way, we get:
  \begin{align}
    \|\vphi_{t+1}-\vphi^*\|^2
    &= \|\vphi - \vphi^*\|^2 + 2\eta (\vtheta_t - \vtheta^*) \mA(\vphi_t - \vphi^*) + 2\eta^2 \|\mA^\top(\vtheta - \vtheta^*)\|^2 \\
    & \overset{\eqref{eq:belong_span}}{\geq}  \|\vphi_t - \vphi^*\|^2 + 2\eta (\vtheta_t - \vtheta^*) \mA(\vphi_t - \vphi^*) + \eta^2 \sigma^2_{\min}(\mA)\| \vtheta_t - \vtheta^*\|^2  \label{ineq:theta_thm1}
\end{align}
Summing \eqref{ineq:phi_thm1} and \eqref{ineq:theta_thm1}, we get
\begin{equation}
    \Delta_{t+1} \geq (1 + \eta^2 \sigma^2_{\min}(\mA)) \Delta_t
\end{equation}
where $ \sigma^2_{\min}(\mA)$ is the minimal (positive) squared singular value of $\mA$.  

Now we can try to handle the case where $\beta_1 = \beta_2 = \beta \neq 0$. 
To prove Thm.~\ref{thm:bilin_sim} we will prove the following Proposition
\begin{proposition}Let $F^{\text{sim}}_{\eta,\beta}$ the operator defined in~\eqref{eq:updates_sim}. 
\begin{itemize}
    \item For $\beta\geq 0$ its radial spectrum is lower bounded by $1 + \eta_1\eta_2 \sigma^2_{\max}(A)$.
    \item For $-1/16 \leq \beta < 0$ its radial spectrum is lower bounded by $1 + \eta_1\eta_2 \sigma^2_{\max}(A)/17$. \label{prop:eigs_sim}
\end{itemize}
\end{proposition}
\proof[\textbf{Proof of Proposition~\ref{prop:eigs_sim}}]
Let us use Proposition~\ref{prop:eigs_bilinear_sim} to get that the eigenvalues of our linear operator are the solutions of 
\begin{equation}
    (x-1)^2(x-\beta)^2 + \eta^2 \lambda x^2 \, , \quad \lambda \in Sp(\mA^\top \mA)\,.
\end{equation}
Let us fix $\lambda >0 $ belonging to $Sp(\mA^\top \mA)$. For simplicity, let us note $\alpha^2 = \eta^2 \lambda$.
We can then notice that this polynomial can be factorized as 
\begin{equation}
    (x-1)^2(x-\beta)^2 + (\alpha x)^2 = \left( (x-1)(x-\beta) + i\alpha x \right) \left( (x-1)(x-\beta) - i\alpha x\right)
\end{equation}
Then the roots of these 2 quadratic polynomials are
\begin{align} \label{eq:z_1}
    z_1&= \frac{1+\beta + i\alpha + ((1+\beta + i\alpha)^2 - 4\beta)^{1/2}}{2} \; , \qquad \;
    z_2 = \frac{1+\beta + i\alpha - ((1+\beta + i\alpha)^2 - 4\beta)^{1/2}}{2} \\
    z_3&= \frac{1+\beta - i\alpha + ((1+\beta - i\alpha)^2 - 4\beta)^{1/2}}{2} \quad \text{and} \quad
    z_4= \frac{1+\beta - i\alpha - ((1+\beta - i\alpha)^2 - 4\beta)^{1/2}}{2} \, .
\end{align}
where $\pm z^{1/2}$ are the complex square roots of $z$ with positive imaginary part. Our goal is going to be to show that $z_1$ has a magnitude larger than $1$. 

We are going to use the fact that 
\begin{equation}
    \Re(z^{1/2}) = \sqrt{\frac{|z|+\Re(z)}{2}} \quad \text{and} \quad 
    \Im(z^{1/2}) =\sqrt{\frac{|z|-\Re(z)}{2}}
\end{equation}
Let us first assume that $\beta <0$.
We have that,
\begin{align}
    \Re(z^{1/2}) 
    &=  \sqrt{\frac{\sqrt{((1-\beta)^2 - \alpha^2)^2 + 4 \alpha^2 (1+\beta)^2}+(1-\beta)^2 - \alpha^2}{2}}\\
    &= \sqrt{\frac{\sqrt{((1-\beta)^2 + \alpha^2)^2 + 16 \alpha^2 \beta}+(1-\beta)^2 - \alpha^2}{2}} \\
    & \geq \sqrt{\frac{ (1-\beta)^2 + \alpha^2 + 16 \frac{\alpha^2 \beta}{\alpha^2 + (1-\beta)^2} +(1-\beta)^2 - \alpha^2}{2}} \\
    & = \sqrt{ (1-\beta)^2 + 8 \frac{\alpha^2 \beta}{\alpha^2+(1-\beta)^2}} \\
    & \geq 1 - \beta + 8 \frac{\alpha^2 \beta}{(1-\beta)(\alpha^2+(1-\beta)^2)} \label{eq:lower_Re}
\end{align}
where for the two inequalities we used $\sqrt{1+x}\geq 1+x \, ,\, \forall x \leq 0$. With the same ideas we can lower bound the Imaginary part of $z^{1/2}$,
\begin{align}
    \Im(z^{1/2}) 
    &=  \sqrt{\frac{\sqrt{((1-\beta)^2 - \alpha^2)^2 + 4 \alpha^2 (1+\beta)^2}-(1-\beta)^2 + \alpha^2}{2}}\\
    &= \sqrt{\frac{\sqrt{((1-\beta)^2 + \alpha^2)^2 + 16 \alpha^2 \beta}-(1-\beta)^2 + \alpha^2}{2}} \\
    & \geq \sqrt{\frac{ (1-\beta)^2 + \alpha^2 + 16 \frac{\alpha^2 \beta}{\alpha^2 + (1-\beta)^2} -(1-\beta)^2 + \alpha^2}{2}} \\
    & = \sqrt{ \alpha^2 + 8 \frac{\alpha^2 \beta}{\alpha^2+(1-\beta)^2}} \\
    & \geq \alpha + 8 \frac{\alpha \beta}{\alpha^2+(1-\beta)^2} \label{eq:lower_Im}
\end{align}
Consequently we can use~\eqref{eq:lower_Re} and~\eqref{eq:lower_Im} to lower bound the magnitude of $z_1$ (defined in Eq.~\ref{eq:z_1}) as,
\begin{align}
        |z_1|^2 
        & = \Re(z_1)^2 + \Im(z_1)^2 \\
        & \geq \left(1  + 4 \frac{\alpha^2 \beta}{(1-\beta)(\alpha^2+(1-\beta)^2)}\right)^2
        + \left(\alpha + 4 \frac{\alpha \beta}{\alpha^2+(1-\beta)^2}\right)^2 \\
        & \geq 1 +  8 \frac{\alpha^2\beta }{\alpha^2+(1-\beta)^2} +\alpha^2 +
        8 \frac{\alpha^2 \beta}{\alpha^2+(1-\beta)^2} \\
        & = 1 + \alpha^2 + 16 \alpha^2\beta
\end{align}
For $-1/16 \leq \beta < 0$ we have that $\alpha^2 + 16 \frac{\alpha^2\beta}{\alpha^2+(1-\beta)^2} \geq \frac{\alpha^2}{17}$. Hence,
\begin{equation}
    |z_1|^2 \geq 1+\frac{\alpha^2}{17}  \, , \quad  \forall -1/16 \leq \beta <0
\end{equation}

Let us now consider the case $\beta \geq 0$. By using the fact that $\sqrt{a+b}\geq \sqrt{a} \,,\, \forall a,b \geq 0$ we have that,
\begin{equation}
    \Re(z^{1/2}) = \sqrt{\frac{\sqrt{((1-\beta)^2 + \alpha^2)^2 + 16 \alpha^2 \beta}+(1-\beta)^2 - \alpha^2}{2}} \geq 1- \beta
\end{equation}
and the same way,
\begin{equation}
     \Im(z^{1/2}) 
    =  \sqrt{\frac{\sqrt{((1-\beta)^2 - \alpha^2)^2 + 4 \alpha^2 (1+\beta)^2}-(1-\beta)^2 + \alpha^2}{2}} \geq \alpha
\end{equation}
I then quickly leads to 
\begin{equation}
    |z_1|^2 \geq 1 + \alpha^2\,.
\end{equation}

\endproof

To conclude this proof we just need to combine Proposition~\ref{prop:eigs_sim} with Lemma~\ref{lemma:rate_eigs} saying that if the spectral radius is strictly larger than 1 then the iterates diverge.
\endproof

\subsection{Proof of Thm.~\ref{thm:bilin_alt}} %
\label{sub:thm:_bilin_alt}
\begin{repproposition}{prop:eigs_bilinear_alt}
\propEigsAlt
Particularly for $\beta_1 = \beta_2 = 0$ and $\eta_1\eta_2 = \eta^2$ we get 
\begin{equation}
    P_{\lambda}(x) = x^2( (x-1)^2 + \eta^2\lambda x^3 \, , \quad \lambda \in \Sp(\mA^\top \mA)
\end{equation}
Giving the following set of eigenvalues,
\begin{equation}
  \{0\} \cup \left\{1 + \eta \frac{-\eta \lambda \pm \sqrt{\eta^2 \lambda^2 - 4 \lambda}}{2}
   \; : \; \lambda \in \Sp(\mA^\top \mA) \right\}
\end{equation}
Particularly for $\beta_1 = -\frac{1}{2}$ and $\beta_2 = 0$ we get 
\begin{equation}
     x [(x-1)^2(x+\frac{1}{2}) + \eta^2\lambda x^2 ] \, , \quad \lambda \in \Sp(\mA^\top \mA)
\end{equation}
\end{repproposition}

\proof[\textbf{proof of Proposition~\ref{prop:eigs_bilinear_alt}}]

Let us recall the definition of $F^{\text{alt}}_{\eta,\beta}$, (for compactness we note $F^{\text{alt}}_{\eta,\beta} = F^{\text{alt}}_{\eta_1,\eta_2,\beta_1,\beta_2}$)
\begin{align}
    F^{\text{alt}}_{\eta,\beta} \begin{bmatrix} \vtheta_t \\\vphi_t \\\vtheta_{t-1} \\ \vphi_{t-1}) \end{bmatrix}
    & \defas \begin{bmatrix}
        \vtheta_t - \eta_1 \mA \vphi_t + \beta_1 (\vtheta_t - \vtheta_{t-1}) \\
        \vphi_t + \eta_2 \mA^\top \vtheta_{t+1}+ \beta_2 (\vphi_t - \vphi_{t-1}) \\
        \vtheta_t \\
        \vphi_t
    \end{bmatrix} \\
   & = \begin{bmatrix}
        \vtheta_t - \eta_1 \mA \vphi_t + \beta_1 (\vtheta_t - \vtheta_{t-1}) \\
        \vphi_t + \eta_2 \mA^\top ( \vtheta_t - \eta_1 \mA \vphi_t + \beta_1 (\vtheta_t - \vtheta_{t-1}))+ \beta_2 (\vphi_t - \vphi_{t-1}) \\
        \vtheta_t \\
        \vphi_t
    \end{bmatrix}
\end{align}
Hence, the matrix $F^{\text{alt}}_{\eta,\beta} $ is,
\begin{equation} \label{eq:prop_equation_matrix}
F^{\text{alt}}_{\eta,\beta} 
 = \begin{bmatrix}
    \begin{matrix}
    (1+\beta_1)\bm{I}_{d} &  -\eta_1 \mA \\
    (1+\beta_1) \eta_2 \mA^\top & (1+\beta_2)\bm{I}_{p} - \eta_1\eta_2 \mA^\top \mA 
    \end{matrix}
    &
    \begin{matrix}- \beta_1 \bm{I}_{d} & \bm{0}_{d,p}  \\
    -\beta_1 \eta_2 \mA^\top& - \beta_2\bm{I}_{p} 
    \end{matrix}\\[4mm]
    \bm{I}_m  & \bigzero_m
    \end{bmatrix} 
\end{equation}
Then the characteristic polynomial of $F^{\text{alt}}_{\eta,\beta}$ is equal to 
\begin{align}
\chi(X) 
    &= \begin{vmatrix}
    (X-1-\beta_1)\bm{I}_{d} &  \eta_1 \mA &  \beta_1 \bm{I}_{d} & \bm{0}_{d,p}  \\
    -(1+\beta_1)\eta_2 \mA^\top & (X-1-\beta_2)\bm{I}_{p} + \eta_1\eta_2\mA^\top \mA & \beta_1 \eta_2 \mA^\top & \beta_2\bm{I}_{d} \\
    -\bm{I}_d & \bm{0}_{d,p} & X\bm{I}_{d}&  \bm{0}_{d,p} \\
    \bm{0}_{p,d} & -\bm{I}_{p} & \bm{0}_{p,d}&  X\bm{I}_p
    \end{vmatrix} \\
    & =  \begin{vmatrix}
    (X-1- \beta_1 + \frac{\beta_1}{X})\bm{I}_{d} &  \eta_1 \mA &  \beta_1 \bm{I}_{d} & \bm{0}_{d,p}  \\
    -(1+\beta_1+\frac{\beta_1}{X})\eta_2 \mA^\top & (X-1-\beta_2+\frac{\beta_2}{X})\bm{I}_{p} + \eta_1\eta_2\mA^\top \mA & \beta_1 \eta_2 \mA^\top & \beta_2\bm{I}_{p} \\
    \bm{0}_d & \bm{0}_{d,p} & X\bm{I}_{d}&  \bm{0}_{d,p} \\
    \bm{0}_{d,p} & \bm{0}_{p} & \bm{0}_{p,d}&  X\bm{I}_p
    \end{vmatrix}
\end{align}
Where in the last line we added the third block column multiplied by $\frac{1}{X}$ to the first one.Then we have
\begin{equation}
\chi(X)  =  \begin{vmatrix}
    (X-1- \beta_1 + \frac{\beta_1}{X})\bm{I}_{d} &  \eta_1 \mA &  \beta_1 \bm{I}_{d} & \bm{0}_{d,p}  \\
    -X\eta_2 \mA^\top & (X-1-\beta_2+\frac{\beta_2}{X})\bm{I}_{p}  & \bm{0}_{p,d} & \beta_2\bm{I}_{p} \\
    \bm{0}_d & \bm{0}_{d,p} & X\bm{I}_{d}&  \bm{0}_{d,p} \\
    \bm{0}_{d,p} & \bm{0}_{p} & \bm{0}_{p,d}&  X\bm{I}_p
    \end{vmatrix}
\end{equation}
where we added to the second block line the first block line by $-\eta_2 \mA^\top$. Then our determinant is triangular by squared blocks of size $m\times m$ and we can write,
\begin{align}
\chi(X)  &= \det(X\bm{I}_m)  \begin{vmatrix}
    (X-1- \beta_1 + \frac{\beta_1}{X})\bm{I}_{d} &  \eta_1 \mA   \\
    -X\eta_2 \mA^\top & (X-1-\beta_2+\frac{\beta_2}{X})\bm{I}_{p}  
    \end{vmatrix}\\
   & =   \begin{vmatrix}
   ( X(X-1- \beta_1) +\beta_1)\bm{I}_{d} & X \eta_1 \mA   \\
    -X^2\eta_2 \mA^\top & (X(X-1-\beta_2)+\beta_2)\bm{I}_{p}  
    \end{vmatrix}\\
   & = \begin{vmatrix}
   (X-1)(X- \beta_1)\bm{I}_{d} & X \eta_1 \mA   \\
    -X^2\eta_2 \mA^\top & (X-1)(X-\beta_2)\bm{I}_{p}  
    \end{vmatrix}\\
    & =  \begin{vmatrix}
   (X-1)(X- \beta_1)\bm{I}_{d} + \eta_1\eta_2 \mA^\top \mA \frac{X^3}{(X-1)(X-\beta_2)} & X \eta_1 \mA   \\
    \bm{0}_{p,d} & (X-1)(X-\beta_2)\bm{I}_{p}  
    \end{vmatrix}
\end{align}
Now we can diagonalize $\mA^\top \mA$  to get,
\begin{align}
  \chi(X)
 = (X-\beta_1)^{d-r}(X-\beta_2)^{p-r}(X-1)^{p+d-2r}  \prod_{k=1}^r \big((X-1)^2(X-\beta_2)(X-\beta_1) + \eta_1\eta_2X^3\lambda_k \big)
\end{align}
where $(\lambda_k)_{1\leq k\leq r}$ are the positive eigenvalues of $\mA^\top \mA$ of rank $r$. 
Particularly, when $\beta_1=\beta_2=0$ we have that,
\begin{equation}
    \chi(X)
    =X^{m} (X-1)^{m-2r} \prod_{k=1}^r \big((X-1)^2+ \eta_1\eta_2X\lambda_k \big)
\end{equation}
\endproof

We report the maximum magnitudes of the eigenvalues of the polynomial from Prop.~\ref{prop:eigs_bilinear_alt} in Fig.~\ref{fig:magnitudes}. 
We observe that they are smaller than 1 for a large choice of step-size and momentum values. 
This is a satisfying numerical result but we want analytical convergence rates.
This is what we prove in Thm.~\ref{thm:bilin_alt}.

\begin{reptheorem}{thm:bilin_alt}
    \thmConvBilin
\end{reptheorem}

\proof 
In Lemma~\ref{lemma:reduc_dim} we showed of the affine transformations $\vtheta_t \rightarrow \mU (\vtheta_t - \vtheta^*)$ and $\vphi_t \rightarrow \mV (\vphi_t - \vphi^*)$ allow us to work on the span of a diagonal matrix $\mD$. Then in that case the eigenspace of $\mD$ do not interact with each other. In the sense that for each coordinate of $[\mU (\vtheta_t - \vtheta^*)]_i$ and $[\mV (\vphi_t - \vphi^*)]_i$ ($1\leq i\leq r$) we have from~\eqref{eq:update_ortho} that 
\begin{equation}
    \left\{
    \begin{aligned}
    [\mU(\vtheta_{t+1}-\vtheta^*)]_i &= [\mU(\vtheta_t-\vtheta^*)]_i - \eta_1 \sigma_i [\mV(\vphi_t-\vphi^*)]_i + \beta_1[\mU(\vtheta_t-\vtheta_{t-1})]_i \\
    [\mV(\vphi_{t+1}-\vphi^*)]_i &= [\mV(\vphi_t-\vphi^*)]_i + \eta_2 \sigma_i [\mU(\vtheta_{t+1}-\vtheta^*)]_i + \beta_2 [\mV (\vphi_t - \vphi_{t-1})]_i
    \end{aligned}
    \right.
\end{equation}
Consequently we only need to study the 4 dimensional linear operators 
\begin{equation}\label{eq:small_matrix}
    \begin{bmatrix}
    \begin{matrix}
    (1+\beta_1) &  -\eta_1 \sigma_i \\
    (1+\beta_1) \eta_2 \sigma_i & (1+\beta_2) - \eta_1\eta_2 \sigma_i^2
    \end{matrix}
    &
    \begin{matrix}- \beta_1  & 0 \\
    -\beta_1 \eta_2 \sigma_i& - \beta_2 
    \end{matrix}\\[4mm]
    \bm{I}_2  & \bigzero_2
    \end{bmatrix} 
\end{equation}
for $\sigma_1\leq \cdots \leq \sigma_r>0$ the positive singular values of $\mA$. These equations are a particular case of~\eqref{eq:prop_equation_matrix}. Using the proof of Proposition~\ref{prop:eigs_bilinear_alt} the eigenvalues of these matrices are the solution of 
\begin{equation}
    P_i(X) = (X-1)^2(X-\beta_1)(X-\beta_2) + \eta_1\eta_2X^3 \sigma_i^2 \, , \quad  1\leq i \leq r\,.
\end{equation}
We will now consider two case:
\begin{itemize}
    \item When, $\beta_1=\beta_2=0$ we have that,
    \begin{equation}
    P_i(X) = X^2((X-1)^2 + \eta_1\eta_2X \sigma_i^2)\, , \quad  1\leq i \leq r\,.
    \end{equation}
    Then the roots of $P_i(X)$ are $0$ and two complex conjugate value with a magnitude equal to the constant term of $(X-1)^2 + \eta_1\eta_2X \sigma_i^2$ which is $1$. 
    Since these two eigenvalues are different, the matrix~\eqref{eq:small_matrix} is diagonalizable 
    (for $\beta_1 = \beta_2=0$ we can remove the state augmentation to only work with these two eigenvector). 
    Consequently our linear operator is diagonalizable and has all its eigenvalues larger than 1 in magnitude, we can then apply Lemma~\ref{lemma:rate_eigs} to conclude that $\Delta_t = \Omega(\Delta_0)$.
    
    \item When, $\beta_1 = -\frac{1}{2}$ and  $\beta_2=0$, we have that $P_i(X) = X Q_i(X)$ where 
    \begin{equation}
     Q_i(X) \defas (X-1)^2(X+\tfrac{1}{2}) + \eta_1\eta_2X^2 \sigma_i^2\, , \quad  1\leq i \leq r\,.
    \end{equation}
    Then $P_i(-1/2) = +\frac{\eta_1\eta_2 \sigma_i^2}{4}>0$ and $P_i(-1)=-2+\eta_1\eta_2 \sigma_i^2$. If $\eta_1\eta_2 < \frac{2}{\sigma_i^2}$ we have $P_i(-1)<0$. Consequently, this polynomial has a negative root $\lambda_-$ such that $-1 < \lambda_-<-\frac{1}{2}<0$. 
    Moreover the derivative of $ Q_i(X)$ is 
    \begin{equation}
     Q_i'(X) = (X-1)(2X+1) + (X-1)^2 + 2\eta_1\eta_2X \sigma_i^2 = (3X-3 - \eta_1\eta_2 \sigma_i^2)X \,.
    \end{equation}
    If $\eta_1\eta_2 < \frac{3}{2\sigma_i^2}$, then $Q'_i(x)>0 \, , \forall x> 0$. Since $Q_i(0)=1/2>0$ then $Q_i(x)>0 \, , \forall x\geq 0$ and consequently all the real roots of $Q_i$ are negative. 

    Since by the root coefficient relationship the sum of the roots of $Q_i$ has to be equal to $\frac{3}{2} - \eta_1\eta_2\sigma_i^2>0$, all the roots of $Q_i$ cannot be real (because the real roots of $Q_i$ are negative). Hence $Q_i$ has two conjugate roots $\lambda_c$ and $\bar \lambda_c$ and one real negative root $\lambda_r$. 
    Let us consider $-1<\lambda_r< -1/2$, we have,
    \begin{equation}
        4(\lambda_r+\tfrac{1}{2}) + \alpha \lambda_r^2 <(\lambda_r-1)^2(\lambda_r+\tfrac{1}{2}) + \alpha \lambda_r^2 = 0 \,, 
    \end{equation}
    where we called $\alpha = \eta_1\eta_2\sigma_i^2$. Thus we have,
    \begin{equation}
         -\frac{2 + \sqrt{4 - 2 \alpha}}{\alpha}<\lambda_r <  \frac{\sqrt{4 - 2 \alpha} - 2}{\alpha} \leq \frac{2 - \alpha/2 - \alpha^2/16  - 2}{\alpha} = -\frac{1}{2} - \frac{\alpha}{16}
    \end{equation}
    where we used $1 - \frac{x}{2} - \frac{x^2}{8} \geq \sqrt{1-x}  \,,\, 1 > x \geq 0$. 
    Moreover the roots coefficient relationship are 
\begin{align}
        \frac{3}{2} - \alpha &= 2 \Re(\lambda_c) + \lambda_r \label{eq:root_coeff_sum} \\
        0 &= |\lambda_c|^2 + 2\lambda_r \Re(\lambda_c) \label{eq:root_coef_crossed}\\
        -\frac{1}{2} &= \lambda_r |\lambda_c|^2 \label{eq:root_coef_product}
\end{align}
    where we called $\alpha = \eta_1\eta_2\sigma_i^2$. 
    Plugging~\eqref{eq:root_coeff_sum} into~\eqref{eq:root_coef_crossed} we get 
    \begin{equation}
        0 = |\lambda_c|^2 + (\tfrac{3}{2} -\alpha-\lambda_r)\lambda_r
    \end{equation}
    Multiplying by $\lambda_r$ and plugging~\eqref{eq:root_coef_product} in we get
    \begin{equation}
        \lambda_r^2 = \frac{1}{3-2\alpha-2\lambda_r} \leq \frac{1}{4-2\alpha}
    \end{equation}
    where we used that $\lambda_r <-\frac{1}{2}$. 
    Consequently, since in the theorem we assumed that $\eta_1\eta_2 \leq \frac{1}{\sigma_{\max}(\mA)^2}$, we have that $\alpha \leq 1$, we have
    \begin{equation}
        \lambda_r^2 \leq \frac{1}{4-2\alpha} \leq \frac{1}{2} \quad \text{and} \quad |\lambda_c|^2 = \frac{-1}{2\lambda_r} \leq \frac{1}{1 + \frac{\alpha}{8} } \leq 1 - \frac{\alpha}{16}
    \end{equation}
    where for the last inequality we used $\sqrt{1+x} \leq 1 + \frac{x}{2} \, , \, \forall x \in \R$ and $(1+x)^{-1} \leq 1 - x/2\,,\, \forall \, 0 \leq x \leq 1.$
    
    One last thing to say is that the four roots of $P_i$ which are the four eigenvalues of the matrix in~\eqref{eq:small_matrix} are different and consequently this matrix is diagonalizable. 
    
    We can then apply Lemma~\ref{lemma:rate_eigs} in a case of a spectral radius strictly smaller that $1$ to conclude that,
    \begin{equation}
       \Delta_{t+1} \leq \max\{1/2,1- \eta_1\eta_2 \tfrac{\sigma^2_{\min}(\mA)}{16}\}\Delta_t
    \end{equation}
    where,
    \begin{align}
    \Delta_t &:=\|\mU(\vtheta_{t+1}-\vtheta^*)\|^2_2 + \|\mU(\vtheta_{t}-\vtheta^*)\|_2^2 +\|\mV(\vphi_{t+1}-\vphi^*)\|_2^2 + \|\mV(\vphi_{t}-\vphi^*)\|^2_2 \\
    &= \|\vtheta_{t+1}-\vtheta^*\|^2_2 + \|\vtheta_{t}-\vtheta^*\|_2^2 +\|\vphi_{t+1}-\vphi^*\|_2^2 + \|\vphi_{t}-\vphi^*\|^2_2        
    \end{align}
 because $\mU$ and $\mV$ are orthogonal.
\end{itemize}

\end{document}